\documentclass[letterpaper]{article} 
\usepackage{aaai2026}  
\usepackage{times}  
\usepackage{helvet}  
\usepackage{courier}  
\usepackage[hyphens]{url}  
\usepackage{graphicx} 
\usepackage{natbib}  
\usepackage{caption} 
\usepackage{algorithm}
\usepackage{amsmath}
\usepackage{mathtools}
\usepackage{todonotes}
\usepackage[export]{adjustbox}
\usepackage{amsfonts}
\usepackage{microtype}
\usepackage{subcaption}
\usepackage{algpseudocode}
\usepackage[capitalize,noabbrev]{cleveref}

\DeclareMathOperator{\defeq}{{\vcentcolon=}}
\newcommand{\E}{\mathbb{E}}
\newcommand{\A}{\mathcal{A}}
\newcommand{\D}{\mathcal{D}}

\providecommand{\shortver}[1]{}
\shortver{\newcommand{\longver}[1]{}}
\providecommand{\longver}[1]{#1}

\usepackage{newfloat}
\usepackage{listings}
\DeclareCaptionStyle{ruled}{labelfont=normalfont,labelsep=colon,strut=off} 
\floatstyle{ruled}
\newfloat{listing}{tb}{lst}{}
\floatname{listing}{Listing}
\title{Practical, Utilitarian Algorithm Configuration}
\author {    
    Devon Graham\textsuperscript{\rm 1},
    Eros Rojas Velez\textsuperscript{\rm 1},
    Kevin Leyton-Brown\textsuperscript{\rm 1}    
}
\affiliations {
    \textsuperscript{\rm 1}Department of Computer Science\\University of British Columbia\\
    drgraham@cs.ubc.ca, erojas20@student.ubc.ca, kevinlb@cs.ubc.ca
}

\begin{document}

\maketitle

\begin{abstract}
Utilitarian algorithm configuration identifies a parameter setting for a given algorithm that maximizes a user's utility. Utility functions offer a theoretically well-grounded approach to optimizing decision-making under uncertainty and are flexible enough to capture a user's preferences over algorithm runtimes (e.g., they can describe a sharp cutoff after which a solution is no longer required, a per-hour cost for compute, or diminishing returns from algorithms that take longer to run). COUP is a recently-introduced utilitarian algorithm configuration procedure which was designed mainly to offer strong theoretical guarantees about the quality of the configuration it returns, with less attention paid to its practical performance. This paper closes that gap, bringing theoretically-grounded, utilitarian algorithm configuration to the point where it is competitive with widely used, heuristic configuration procedures that offer no performance guarantees. We present a series of improvements to COUP that improve its empirical performance without degrading its theoretical guarantees and demonstrate their benefit experimentally. Using a case study, we also illustrate ways of exploring the robustness of a given solution to the algorithm selection problem to variations in the utility function.
\end{abstract}

\begin{links}
    \link{Code}{https://github.com/drgrhm/practical-coup}
\end{links}

\section{Introduction}

Algorithm configuration is the process of finding a complete setting of an algorithm's parameters that lead it to perform well on a given set of input instances. Often, the algorithm of interest is a solver for a hard computational problem such as Boolean Satisfiability (SAT). Since SAT is NP-complete, we are unlikely to be able to find a single solver or configuration that performs well on all possible inputs. In practice, however, we are never interested in solving \emph{all possible} inputs. Instead, we  tend to be interested in solving input instances drawn from a given distribution. 

Algorithm configuration is thus a  bandit problem \citep{lattimore2020bandit} where each configuration is an arm, and running a configuration on an input instance means drawing a sample from that arm's underlying distribution. Unlike the standard bandit setting, in which there is a unit cost associated with pulling an arm, in algorithm configuration the cost is the time it takes for the algorithm to solve the input. We can also choose whether each such evaluation should be censored (``capped'') at some maximum runtime. The objective in algorithm configuration is not regret minimization, but rather (approximate) best arm identification.

Algorithm configuration procedures have traditionally aimed to minimize the expected runtime of the returned configuration. The most practical such methods make these decisions according to carefully chosen heuristics \citep{birattari2002racing,hutter2009paramils,ansotegui2009gender,hutter2011sequential,lopez2016irace,SMAC3}. A second line of work has identified (considerably less practical) algorithms that offer worst-case  guarantees about the runtime required to identify approximately optimal solutions \citep{kleinberg2017efficiency,weisz2018leapsandbounds,kleinberg2019procrastinating,weisz2019capsandruns,weisz2020impatientcapsandruns}. For the purposes of this paper, we define an optimality guarantee as depending on three parameters: $\epsilon$, $\gamma$, and $\delta$. With failure probability $\delta$, we want to prove both that the returned configuration is within an additive factor of $\epsilon$ from the best configuration sampled, and that the probability of sampling a configuration that is better than the best one sampled so far is at most $\gamma$.

It has recently been argued that algorithm configuration should rest on the firmer decision-theoretic foundation of maximizing a utility function over algorithm runtime \citep{graham2023formalizing}. That is, let $u$ be a function that maps runtimes to utility values in $[0, 1]$, encoding the user's preferences about runtimes of different lengths; we should then maximize $u$'s expected value. A recent line of work has proposed procedures that offer worst-case guarantees about the runtime required to approximately optimize a utilitarian objective \citep{graham2023utilitarian,brandt2023ac}. The most recent such procedure is COUP (Continuous, Optimistic Utilitarian Procrastination; \citealt{graham2024utilitarian}), which was shown empirically to perform favorably with respect to other configuration procedures that offer (in some cases, somewhat incomparable) optimality guarantees: Successive Halving \citep{jamieson2016non}, Hyperband \citep{li2018hyperband}, AC-Band \citep{brandt2023ac}, ImpatientCapsAndRuns \citep{weisz2020impatientcapsandruns} and Structured Procrastination with Confidence \citep{kleinberg2019procrastinating}. Pseudocode for COUP is included in \cref{app:code}\shortver{ of the extended version of the paper}, but in essence it is an instantiation of the UCB algorithm \citep{lai1985asymptotically,lai1987adaptive}, coupled with a captime exploration technique and a condition for expanding the set of configurations under consideration. For each configuration $i$, COUP maintains upper and lower confidence bounds $UCB_i$ and $LCB_i$. In each iteration it chooses the configuration $i' = \arg\max_i UCB_i$ and performs a run of it. In this way it works to improve $\epsilon^* = \max_i UCB_i - \max_i LCB_i$, which is the best optimality guarantee it can make. If COUP detects that the captime it is using is too small, it will double it. If COUP detects that its set of configurations is too small, it will add more. COUP is \emph{anytime} in its guarantees with respect to both $\epsilon$ and $\gamma$: as it runs it makes tighter and tighter optimality guarantees with respect to larger and larger sets of configurations. COUP is also \emph{adaptive} to the inputs it is given: it provably makes better guarantees in less time when given easier inputs. 

This paper begins in \cref{sec:improvements} by describing a series of improvements to COUP that boost its empirical performance without compromising its theoretical guarantees. First, we refine the confidence bounds used by COUP, making the optimization process more efficient. Second, we incorporate a different version of the UCB algorithm that specifically targets the best arm identification problem. Third, we suggest a new condition for when COUP should add new configurations, which frees the user from having to provide a set of parameters ahead of time, instead adapting to the inputs it sees. And fourth, we make COUP model-based, using a boosted forest of regression trees to help COUP find new configurations. In \cref{sec:experiment} we show for the first time that an algorithm configuration method offering theoretical guarantees (our modified version of COUP) robustly achieves performance roughly equivalent to that of SMAC 
\citep{hutter2011sequential,SMAC3}, a heuristic procedure widely seen as the empirical state of the art. 
Finally, in \cref{sec:utility_functions}, we address a shortcoming of the utilitarian framework: that it is not always easy for the end user to choose the exact utility function to optimize. Via a case study on data from one of the most prominent international evaluations of heuristic algorithms, the SAT Competition, we illustrate how runtime data can be used to reason about the utility functions for which a given algorithm would be considered optimal.

\section{Practical Improvements to COUP}\label{sec:improvements}

We now present four of practical improvements to COUP that significantly improve its performance, both in terms of the optimality guarantees it is able to make and in terms of the empirical quality of the configuration it returns. \cref{sec:klbound} discusses a tightening of the confidence bounds used by COUP, \cref{sec:lucb} argues for using a variant of the UCB algorithm, \cref{sec:newconfigs} discusses an improvement to the way COUP incorporates new configurations during its search process, and \cref{sec:model} discusses using a model to help guide COUP's search for new configurations.

\subsection{Improved Confidence Bounds}\label{sec:klbound}

A key technical tool used in COUP is Hoeffding's Inequality \citep{hoeffding1963pobability}. This concentration inequality asserts that for bounded random variables, the probability that the observed sample mean is far from the true population mean is exponentially small in the number of samples taken. However, Hoeffding's Inequality involves a key approximation, which is tight when the true mean is close to $1 / 2$ (in the case of random variables bounded on [0,1]), but grows worse when the true mean is close to either 0 or 1---a situation encountered often in utilitarian algorithm configuration. We can do away with this approximation by appealing directly to the ``Chernoff--Hoeffding Lemma'', and solving numerically for the confidence bounds, which are then guaranteed to be tighter than a naive application of Hoeffding's Inequality; for a detailed exposition see \citet{lattimore2020bandit}, Chapter 10. The confidence bounds come from solving an optimization problem that bounds the KL-Divergence between the observed sample mean and the bounding quantity. We thus refer to them as ``KL bounds''. 

COUP maintains a captime $\kappa$, which is the maximum time it will spend on each run of a configuration. To derive upper and lower confidence bounds for a configuration $i$'s mean utility, COUP needs to estimate its capped mean runtime $U_i = \mathbb{E}_{t_i}[u(\min(t_i, \kappa))]$ and its probability of not capping $F_i = \Pr_{t_i}(t_i \le \kappa)$. Both of these estimates can be improved by switching from Hoeffding bounds to KL bounds. Define $d(p, q) = p \log(p / q) + (1 - p) \log ((1 - p) / (1 - q))$.
Suppose $n$ is the current number of configurations, $m$ is the number of samples we have taken for configuration $i$, and $\kappa$ is the captime we have taken them at. If $\widehat{U}_i$ is the empirical capped average runtime, $\widehat{F}_i$ is the empirical fraction of completed runs, and $a = \frac{1}{m}\ln\frac{27 n^2 m^2 (\log(\kappa + 1)^2}{\delta}$ then 
\begin{align*}
    UCB_i &= \max\big\{ u \in  [0, 1] \,:\; d(\widehat{U}_i, u) \le a \big\}, \text{ and} \\
    LCB_i &= \min\big\{ u \in  [0, 1] \,:\; d(\widehat{U}_i, u) \le a \big\} \\
    &\quad - u(\kappa) \big(1 - \min\big\{ u \in  [0, 1] \,:\; d(\widehat{F}_i, u) \le a \big\}\big)
\end{align*}
are confidence bounds that hold for all configurations at all times during COUP's execution with probability at least $1 - \delta$ (detailed derivations can be found in \cref{app:confbounds}\shortver{ of the extended version of the paper}). This alteration essentially makes COUP an implementation of the KL-UCB algorithm \citep{cappe2013kullback} for bandit problems. In \cref{fig:epsilons} (Left) in \cref{sec:experiment} we show that COUP performs better when using these refined bounds. 
 
\subsection{Lower Upper Confidence Bound}\label{sec:lucb}

The UCB algorithm is a natural choice as a framework for algorithm configuration procedures. Indeed, \citet{graham2024utilitarian} argue that UCB is the best way to apply utilitarian algorithm configuration to infinite parameter spaces, and show that using UCB is a significant improvement over configuration procedures that implement other bandit procedures. However, UCB is essentially designed for regret minimization, while the algorithm configuration problem is a case of best arm identification. Thankfully, the LUCB (\emph{Lower} Upper Confidence Bound) algorithm \citep{gabillon2012best,kalyanakrishnan2012pac} is a variant of UCB that was designed to be applied to the best arm identification problem. Actually, the LUCB algorithm was designed to solve the $m$-best arm identification problem, where a learner is tasked with identifying the top $m$ arms from a set of $n$ arms. Although we are only concerned with the special case of $m=1$, it is easier to understand what LUCB is doing by considering the more general case. In each iteration, LUCB selects and samples \emph{two} arms (i.e., configurations). First it samples the arm with the smallest lower confidence bound from the $m$ empirically best arms. Next it samples the arm with the largest upper confidence bound from among the $n - m$ remaining arms. This way, the LUCB algorithm attempts to tease apart the $m$ best arms from the $n - m$ worst arms. \cref{fig:epsilons} (Left) in \cref{sec:experiment} shows an example of the improvement achieved by using the LUCB algorithm. This is drastic early in a search.

\subsection{Adding New Configurations}\label{sec:newconfigs}

COUP runs in phases. In each phase it reduces both $\epsilon$ and $\gamma$ by an amount that is specified by the user at runtime. This approach is non-adaptive; it requires that the user specify a sequence of $\epsilon$ and $\gamma$ values as parameters, which is not only burdensome but also does not automatically take into account any properties of the actual runtime data COUP sees. For instance, if COUP finds that the best utility it can prove is relatively low, then it should tend to favor sampling new configurations more than improving the optimality guarantee of existing configurations. To this end, we define two conditions for adding new configurations that adapt to the relative sizes of the configurations' confidence bounds. 

The first condition simply balances the $\epsilon$ and $\gamma$ that COUP can prove. Defining $i^*= \arg\max_i LCB_i$ and $i' = \arg\max UCB_i$, if COUP adds new configurations when $UCB_{i'} - LCB_{i^*} < \gamma$ then it will work to reduce the $\epsilon$ it can guarantee, until this is smaller than the $\gamma$ it can guarantee, at which point it adds a new configuration. This frees the user from having to specify a sequence of $\epsilon$ values, but does not have the property that it explores more new configurations when the best utility it can prove is relatively low. 

\begin{figure}
    \centering
    \includegraphics[width=0.7\linewidth]{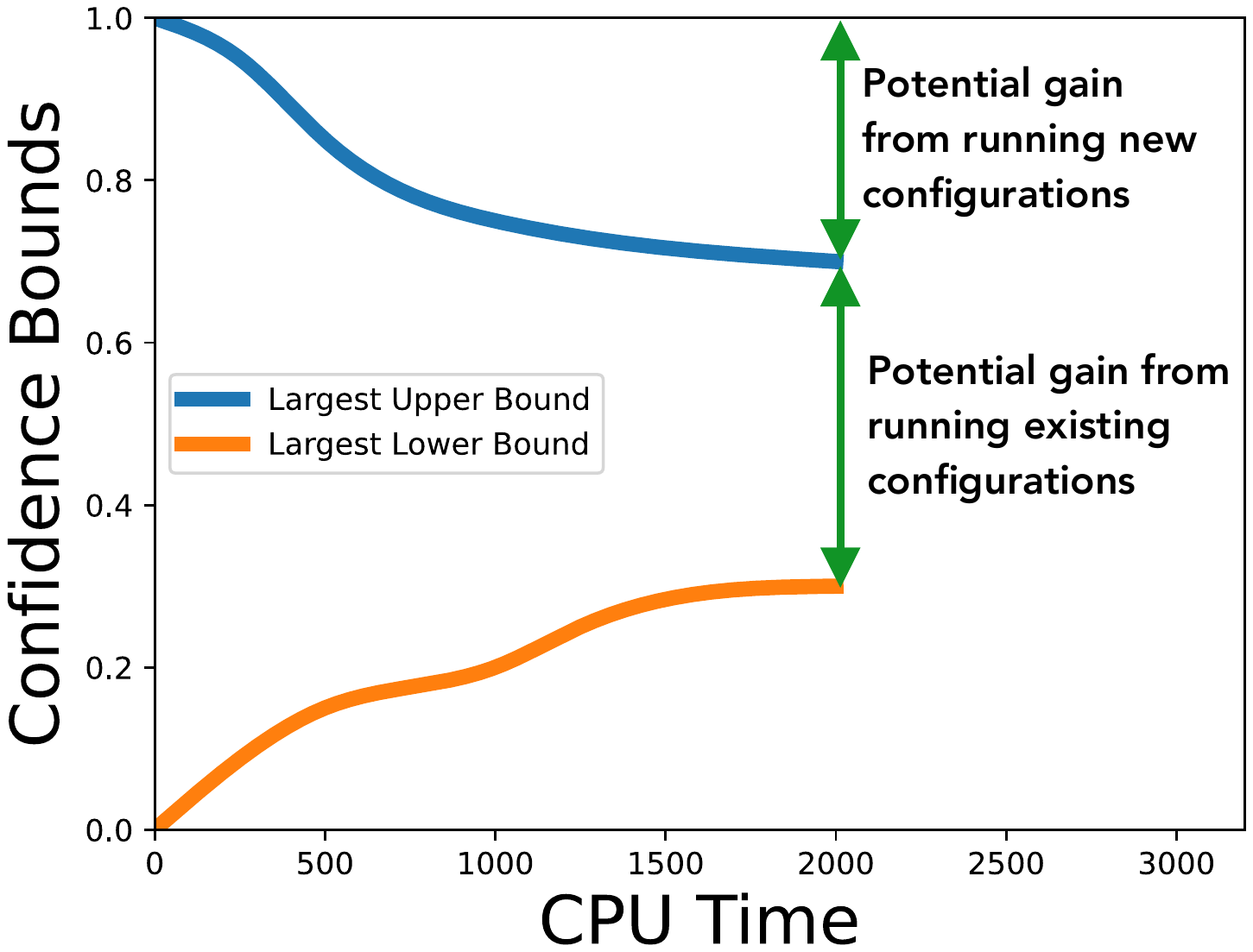}
    \vspace{-.5cm}
    \caption{Illustration of the decomposition that leads to our adaptive strategy for incorporating new configurations.}
    \label{fig:maxbounds}
\end{figure}

The second condition does have this property. At any point during its execution, COUP will recommend the configuration $i^* = \arg\max LCB_i$, the configuration with largest lower confidence bound. This is the configuration about which it can prove the best guarantee. The quantity $LCB_{i^*}$ is the highest expected utility that COUP can guarantee at any time. We would like to push $LCB_{i^*}$ as close to 1 as possible because then we will have (i) a very good configuration, and (ii) a very strong guarantee about this configuration. So we can think of the quantity $1 - LCB_{i^*}$ as the utility ``still on the table''; the largest amount of utility by which we might hope to improve our guarantee. We can write $1 - LCB_{i^*} = 1 - UCB_{i'} + UCB_{i'} - LCB_{i^*}$. The term $UCB_{i'} - LCB_{i^*}$ is the best $\epsilon$ we can guarantee with respect to our existing set of configurations; the maximum amount by which we can shrink $1 - LCB_{i^*}$ by running existing configurations. On the other hand, since $UCB_{i'}$ is the largest upper bound from our current sample of configurations, we can be confident that the configurations we have seen so far have utility at most $UCB_{i'}$. So the term $1 - UCB_{i'}$ is the utility potentially to be gained from other configurations that we have not yet seen. This is illustrated in \cref{fig:maxbounds}.

The magnitude of $UCB_{i'} - LCB_{i^*}$ tells us what is potentially to be gained by running our existing configurations, while the magnitude of $1 - UCB_{i'}$ tells us what is potentially to be gained by sampling and running new configurations. This suggests a simple condition: to add new configurations when $UCB_{i'} - LCB_{i^*} < 1 - UCB_{i'}$, which means to add new configurations when there is more to be gained from doing so than not. However, with this condition there is a slight problem. If we add new configurations when $UCB_{i'} - LCB_{i^*} < 1 - UCB_{i'}$, then while COUP is running existing configurations it has $UCB_{i'} - LCB_{i^*} \ge 1 - UCB_{i'}$. This means that we will always have $UCB_{i'} \ge \frac{1 + LCB_{i^*}}{2}$ and so $UCB_{i'} - LCB_{i^*} \ge \frac{1 - LCB_{i^*}}{2} \ge \frac{1 - U_{i^*}}{2}$. This means that the $\epsilon$ we can guarantee will never approach 0 but will asymptote (unless $U_{i^*} = 1$, in which case we will have $UCB_{i^*} = 1$ also and we will not sample any more configurations, which is what we want since $U_{i^*}$ is already maximal).

Both conditions have downsides but by combining them together we get a condition that avoids these drawbacks. If we add new configurations when $(UCB_{i'} - LCB_{i^*})^2 < \gamma(1 - UCB_{i'})$, then we will tend to add new configurations when $1 - UCB_{i'}$ is large, but then this will cause $\gamma$ to be reduced and COUP will tend to make a corresponding reduction to $\epsilon^* = UCB_{i'} - LCB_{i^*}$. Thus this condition will tend to continue to shrink $\epsilon$ and also is adaptive to the inputs given. We note that this condition is equivalent to comparing $\epsilon^*$ on the left to the geometric mean of $\gamma$ and $1 - UCB_{i'}$ on the right.

\subsection{Model-guided Configuration Search}\label{sec:model}

COUP can be improved by incorporating a learning model that predicts the performance of candidate configurations before spending significant amounts of time investigating them. If good configurations are rare enough, then we will be all but guaranteed to never encounter them by random sampling alone. A model will at least have the possibility of finding such configurations, and there is evidence for this: \citet{hutter2011sequential} report that SMAC outperforms ROAR, its model-free variant, in all scenarios and substantially outperforms ROAR on some large configuration spaces, indicating that at least in some cases the prediction model is able to find significantly better configurations than random sampling alone. 

As COUP runs, it generates the performance data needed to train a performance model and we can use the trained model to help identify promising areas of the search space to investigate. Areas that are predicted to yield configurations with high performance will be investigated more, as will areas that have high uncertainty. We use a gradient-boosted set of regression trees as our base learning model to predict configuration performance. We use the XGBoost \citep{chen2016xgboost} implementation for gradient boosting. In order to produce uncertainty estimates we perform a bootstrap aggregation procedure, sampling  subsets of configurations and training an XGBoost model on each one. For each candidate configuration we make a prediction with each of the models, giving us a distribution of values. We then consider 95\% confidence intervals, giving us an uncertainty band around each prediction. See \cref{app:model}\shortver{ in the extended version of the paper} for more details.

Given a trained model, we employ the same procedure used by SMAC \citep{hutter2011sequential} to identify promising configurations to investigate. Briefly, we start with the ten best configurations seen so far and perform a local search starting with each. For discrete parameters, we consider the set of all configurations that differ in exactly one parameter value. For continuous parameters, we normalize values to the range $[0, 1]$ and sample four neighbors according to a truncated Gaussian distribution with mean equal to the parameter value and standard deviation equal to 0.2. We continue this search as long as the neighbors have a larger upper confidence bound. Once we have sampled 10 neighbors, we use the model to predict their performance, as well as the performance of 10,000 randomly selected configurations, choosing the one that has the highest predicted 95\% upper confidence bound as the next candidate. 

The parameter $\gamma$ described above indicates how likely it is that we sample a configuration that is better than any we have seen so far. As we drive $\gamma$ to 0, we grow more and more sure that we are unlikely to find a better configuration than our current incumbent. However, making this proof requires a growing number of randomly sampled configurations. To continue making this guarantee, we only sample every second configuration using the model, with the other half of the samples chosen randomly. In this case we have a sample of $n' = n_0 + \lfloor \frac{n - n_0}{2} \rfloor \ge \frac{n}{2}$ random configurations, where $n_0$ is the number of initial configurations considered, before the model is trained. Thus, we need to sample twice as many configurations to make the same $\gamma$ guarantee, but this is just a (small) constant, and $\gamma$ is still driven toward 0 as $n \to \infty$ (see \cref{app:gamma}\shortver{ of the paper's extended version} for a full derivation). We retrain the model before asking it to suggest configurations, thus incorporating the new training data we have observed. In \cref{sec:experiment} we show empirically that using a prediction model helped COUP find good configurations more quickly, and that in general configurations found by the model tended to be better than the configurations found by random sampling.

\section{COUP's Empirical Performance}\label{sec:experiment}

\begin{figure*}
    \centering
    \includegraphics[height=11em]{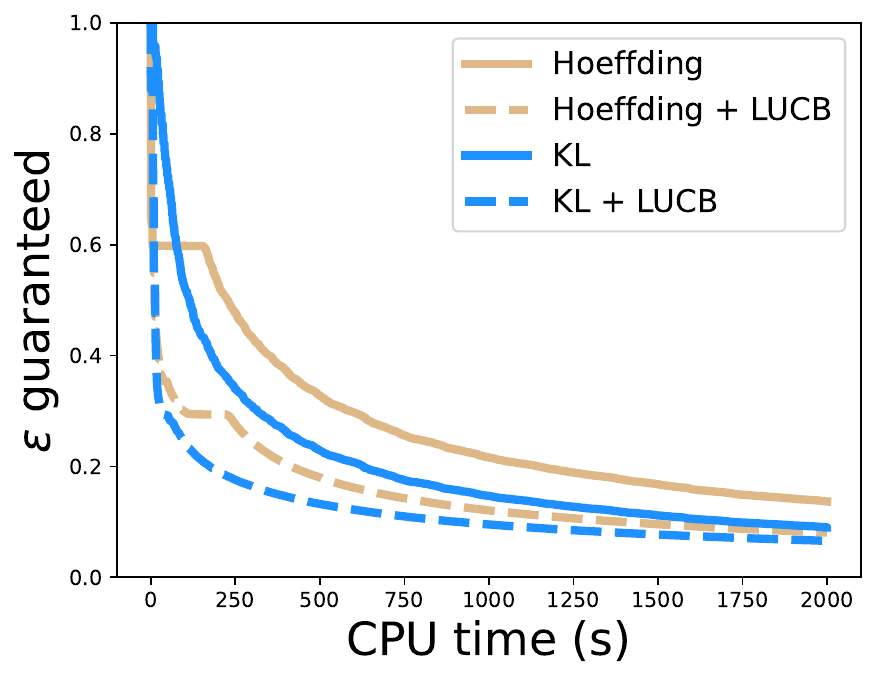}\hfill
    \includegraphics[height=11em]{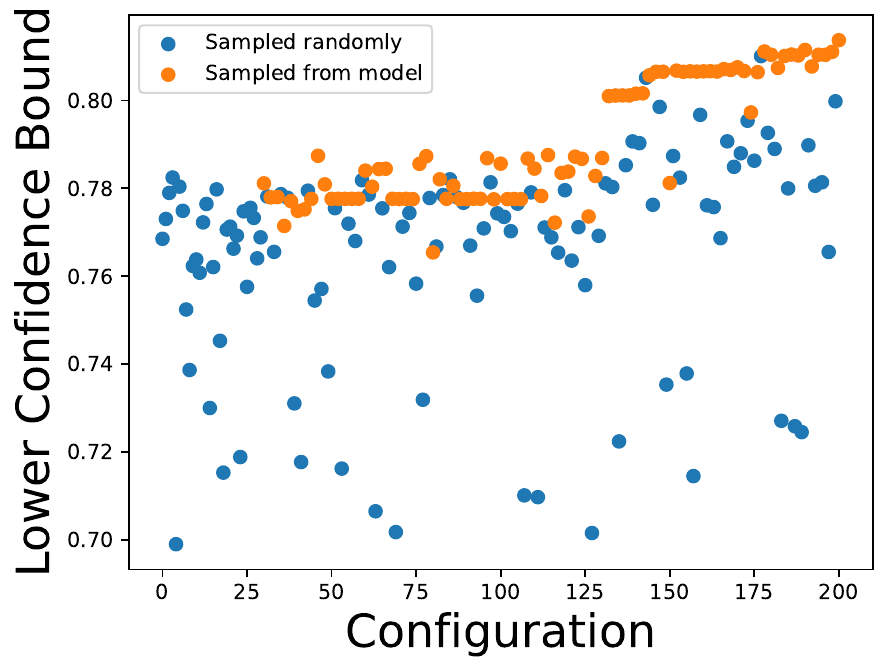}\hfill
    \includegraphics[height=11em]{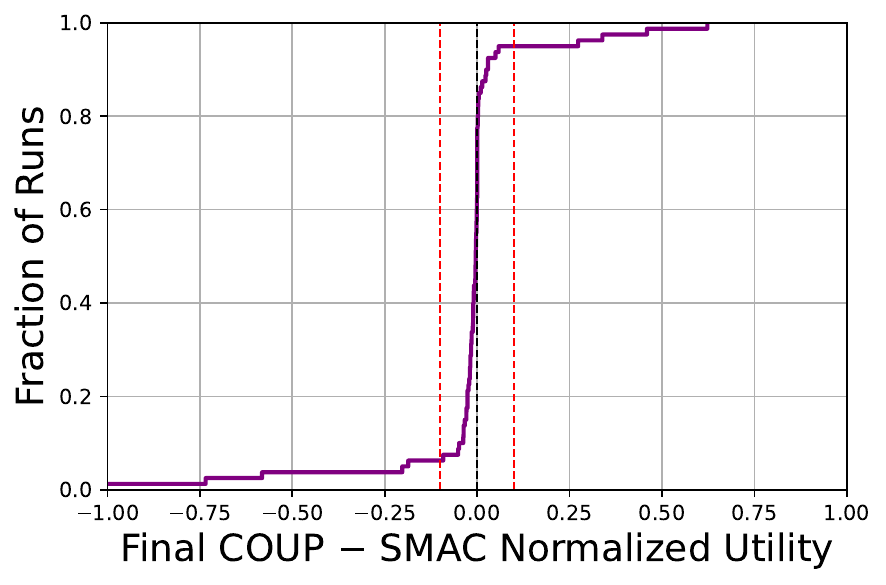}
    \vspace{-.2cm}
    \caption{Left: Suboptimality guarantee provided by COUP, demonstrating the improvements achieved by using the KL bounds and the LUCB algorithm. Center: Lower confidence bounds for configurations sampled by COUP. Configurations sampled according to the model tend to be provably better and to improve with time as more training data is gathered. Right: The CDF of the difference in normalized final utility between COUP and SMAC. Most of the time, COUP and SMAC perform equivalently.}
    \label{fig:epsilons}
    \label{fig:model_configs}
    \label{fig:percent}
\end{figure*}

We have argued theoretically, and provide experimental evidence below, that our changes from Section 2 only improved COUP, first by tightening the bounds and refining the implementation of UCB (\cref{fig:epsilons}, Left) and second by incorporating a performance model (\cref{fig:model_configs}, Center). We also compare COUP to the heuristic procedure that is most widely used in practice: SMAC. While SMAC was initially proposed for the optimization of expected runtime, it is able to optimize any objective function, and so can be straightforwardly extended to a utilitarian objective. We thus compared COUP's performance to that of SMAC 3 \citep{SMAC3} in scenarios consisting of four different SAT solvers and 20 different instance sets. 

We perform experiments using the ACLib library of algorithm configuration benchmarks \citep{hutter2014aclib}. ACLib is well-established as a benchmark set for demonstrating algorithm configuration performance. To the best of our knowledge, ours is the first paper to demonstrate utilitarian algorithm configuration with general optimality guarantees in an online setting, using real runs from real solvers. In contrast, previous work has used precomputed or synthetic data. 

We ran experiments on a 14-node CPU cluster. Each node consisted of 32 dual-socket Intel Xeon E5-2683 CPUs with 96GB of memory. Each COUP scenario (solver and dataset combination) was trained for 96 hours with 12GB of memory per job, with each validation instance having a timeout of 60 seconds once training had been completed. We use the \emph{runsolver} timing tool \citep{roussel2011controlling} to obtain runtime measurements that accurately reflected the total CPU time consumed by each solver. 

For each solver--instance set pair, we evaluated performance on a random sample of 2000 training instances (unless the dataset consisted of fewer than 2000 instances). After each iteration of COUP's training, we saved the best configuration and validated each unique configuration found. We then evaluated SMAC's incumbent configuration at the same points in time and on the same set of validation instances. We limited the individual solver runs that SMAC performed to 60 seconds and used the same limit when validating both COUP and SMAC. We used an initial captime of 1 second for COUP, which is the smallest execution time that \emph{runsolver} is able to enforce. We set COUP's failure probability $\delta = 0.01$. 

The utility function we used has two parameters: $\kappa_0$ and $\kappa_1$. If $t < \kappa_0$ then $u(t) = 1$, indicating that all runs faster than $\kappa_0$ are identical to us and essentially instantaneous. If $t > \kappa_1$ then $u(t) = 0$, indicating that all runs longer than $\kappa_1$ are essentially worthless to us. Between these two extremes, when $t \in [\kappa_0, \kappa_1]$ we have $u(t) = \frac{\log(t / \kappa_1)}{\log(\kappa_0 / \kappa_1)}$, meaning that if runtime is doubled then utility is reduced by a constant factor for any runtime in $[\kappa_0, \kappa_1]$. Optimizing this utility function, which amounts to a scaled and shifted logarithmic transform of runtime, will return configurations that are good for runtimes that are of different orders of magnitude, making it applicable over a broad range of problems. Moreover, using a logarithmic transform of runtime has been shown to improve the quality of algorithm runtime prediction models \citep{xu2008satzilla}. We set $\kappa_0 = 0.001$ seconds, which is approximately the time it takes the operating system to launch a process and thus is the absolute minimum runtime we could observe. We set $\kappa_1 = 1$ hour, which is the maximum amount of time we imagine waiting for any one run to finish. Of course other utility functions are possible. We note that some utility functions might be more ``strict'' than others and will take more time to optimize. In general, the utility functions we consider drop from 1 to 0 as runtime increases. If this drop tends to happen at large runtime values, then COUP will tend to have to do longer runs, and take more time (depending on the given runtime distributions). 

In \cref{sec:klbound} we argued that using the KL bounds instead of naive Hoeffding bounds means that COUP is able to prove a better $\epsilon$ more quickly. In section \cref{sec:lucb} we argued that using the LUCB algorithm rather than the vanilla UCB algorithm also means COUP proves better guarantees more quickly. \cref{fig:epsilons} (Left) shows empirically that both of these changes improve the $\epsilon$ that COUP is able to prove. This plot shows results for 100 configurations of the Spear SAT solver on the FACTORING instance set from ACLib. Other scenarios show similar results. We have tended to observe that using the improved bound helped the most in terms of returning a configuration with good utility, while the use of the LUCB algorithm helped the most in terms of guaranteeing a good $\epsilon$.

In \cref{sec:model} we argued that using a prediction model would help COUP find better configurations overall. \cref{fig:model_configs} (Center) evaluates this process. We first sampled 30 configurations at random, then  alternated between sampling at random and sampling according to the model. Configurations with a higher index ($x$-axis) were sampled later than configurations with a lower index. Not only did configurations sampled according to the model tend to be better than configurations sampled at random, but we can clearly see that the quality of the configurations suggested by the model improved with time, as the model received more training data. The plot shows the performance of the Spear solver on the FACTORING instance set from ACLib. Other scenarios showed qualitatively similar results.

Our augmented version of COUP identified good configurations about as quickly as SMAC at our cutoff time of four wall-clock days. For every (solver, instance) pair, we define the normalized utility as the utility achieved by a given algorithm configuration run divided by the best such utility observed across both the SMAC and COUP runs. \cref{fig:percent} (Right) is an empirical CDF of the difference between the normalized utilities achieved by COUP and SMAC across the different solver and instance set pairs. We call runs that achieved solutions within 10\% of each other effectively equivalent, represented as the region between the dashed red lines. $71$ out of our $80$ runs fell within this region; of what remains, $5$ had a difference less than $-0.1$ (meaning SMAC outperformed), and $4$ had a difference greater than $0.1$ (COUP outperformed). We performed the same analysis at various earlier cutoff times and observed qualitatively similar results.

\begin{figure*}
\includegraphics[width=0.45\textwidth]{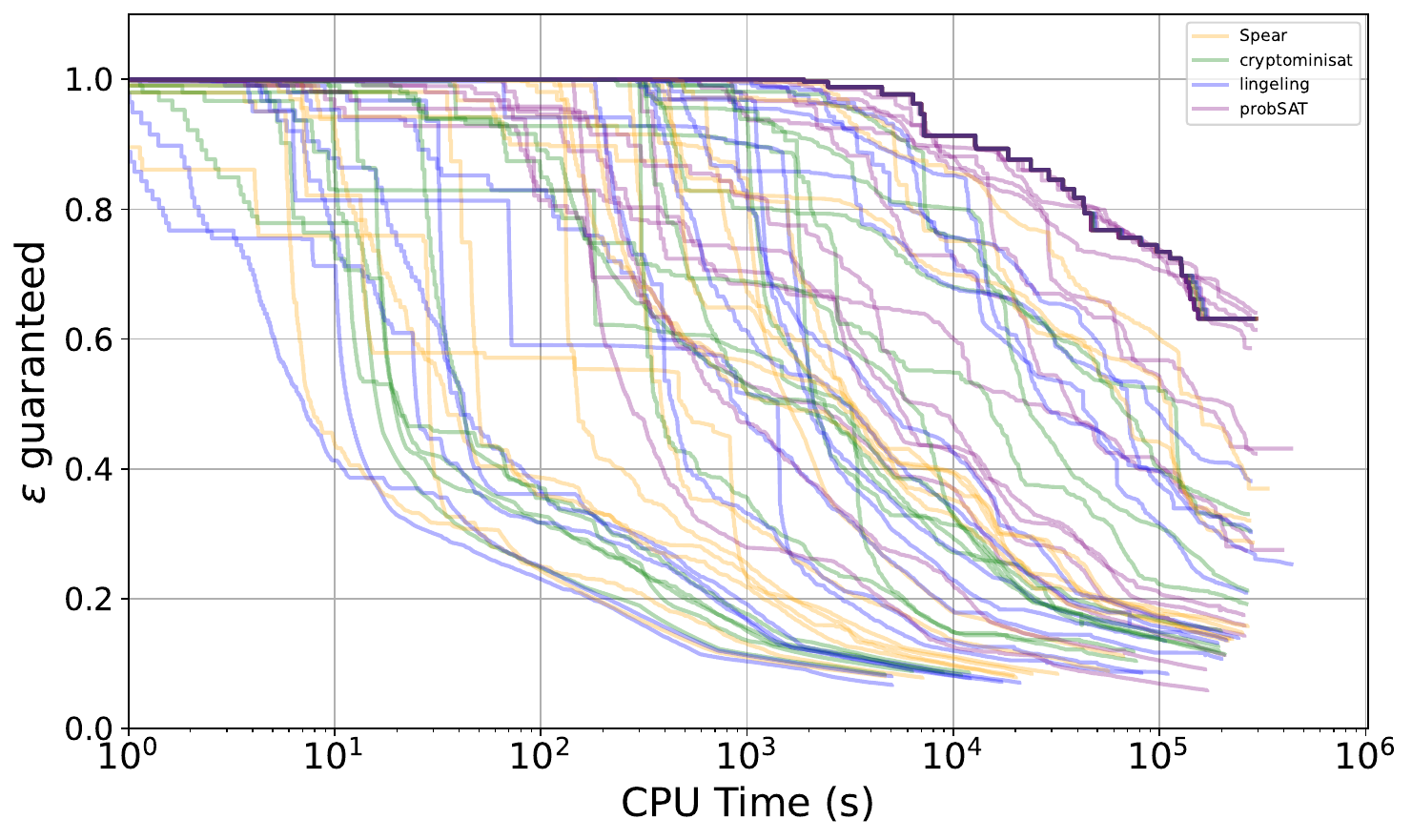}\hfill%
\includegraphics[width=0.45\textwidth]
{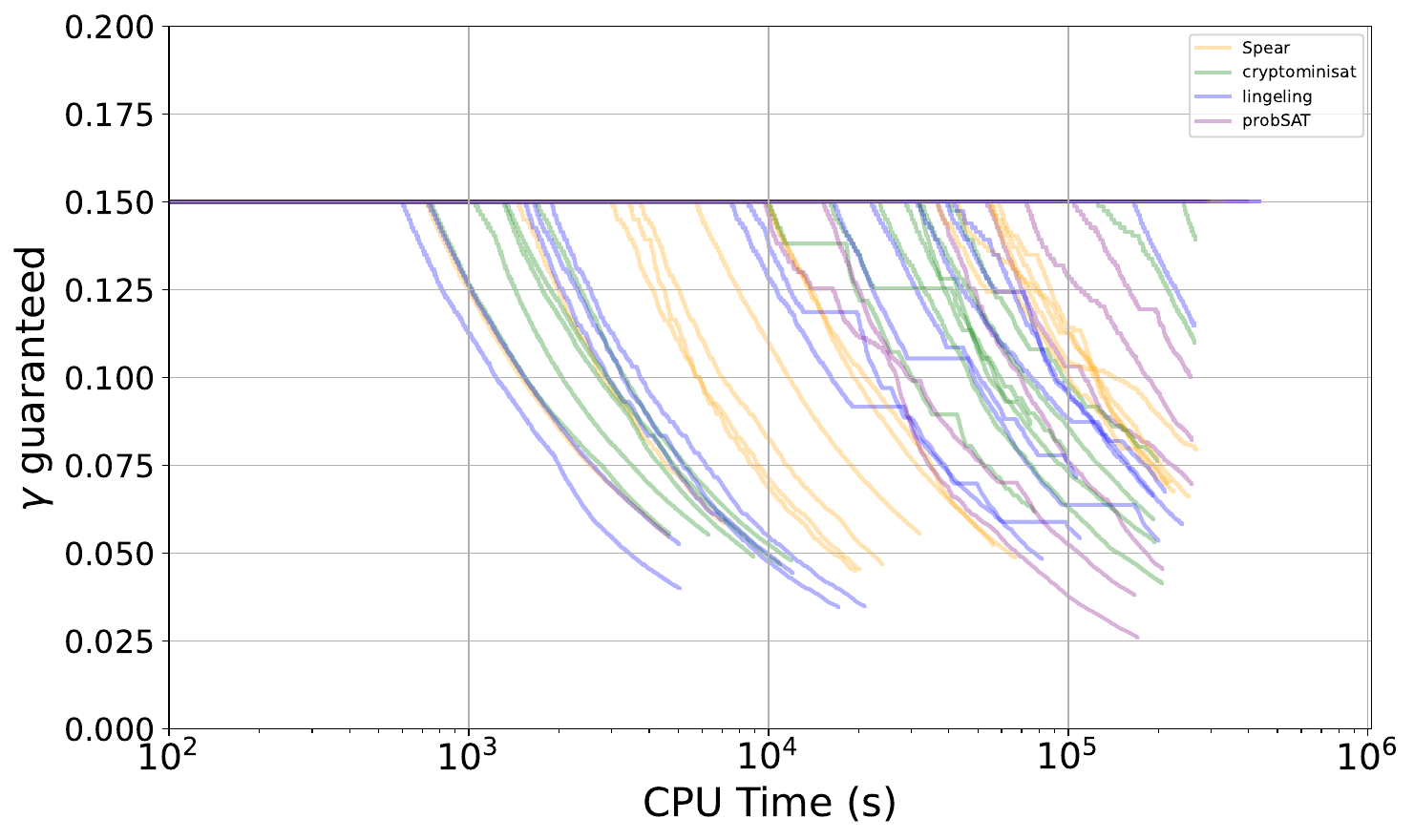}
\vspace{-.2cm}
\caption{Unlike SMAC, COUP also offers theoretical guarantees. In general, both $\epsilon$ and $\gamma$ are driven toward 0.}
\label{fig:epsilon_plot}
\label{fig:gamma_plot}
\end{figure*}

It is worth noting that COUP solves a considerably harder problem than SMAC. Rather than simply identifying strong algorithm configurations, it also proves anytime theoretical guarantees about the quality of the solution it has found at every point in time, both in terms of the amount by which the returned configuration might be worse than some other configuration under consideration ($\epsilon$) and in terms of the probability that a better configuration could be sampled ($\gamma$). 
\cref{fig:epsilon_plot} shows the optimality guarantees proved by COUP for different solver--instance set pairs. Generally, both $\epsilon$ (Left) and $\gamma$ (Right) are driven toward 0, but some trends are evident. ProbSAT struggles to prove a good $\epsilon$ (purple curves tend to be higher). All solvers struggle to prove a good $\epsilon$ on at least one dataset (dark cluster at the top of the plot), and could use with more training time to push $\epsilon$ even farther down. \cref{fig:epsilon_plot} (Right) begins with a solid line (hence the compressed $x$ axis) because COUP tightens only $\epsilon$ until it has learned sufficiently about its initial set of 100 configurations. Plots for individual runs can be found in \cref{app:plots}\shortver{ of the extended version of the paper}. 



Finally, \citet{graham2024utilitarian} included a comparison of basic COUP to SuccessiveHalving \citep{jamieson2016non}, Hyperband \citep{li2018hyperband}, AC-Band \citep{brandt2023ac}, Structured Procrastination \cite{kleinberg2019procrastinating}, and ImpatientCapsAndRuns \citep{weisz2020impatientcapsandruns}. All of these procedures make theoretical guarantees of some sort, but it is important to note that these guarantees are not necessarily comparable to one another. COUP was found to perform well compared to all of these competitors, as was its finite-configuration-set variant, OUP. We included our improved variant of COUP (and OUP) in this comparison. \cref{fig:pgap_plots} in \cref{app:pgap}\shortver{ of the paper's extended version} plots the gap between the best configuration found and the best configuration overall for each of these procedures. It can be seen that our improved variants perform even better than their basic counterparts, in some cases quite significantly.

\section{Utility Functions from Runtime CDFs}\label{sec:utility_functions}

\begin{figure}
    \centering
    \includegraphics[width=.6\linewidth]{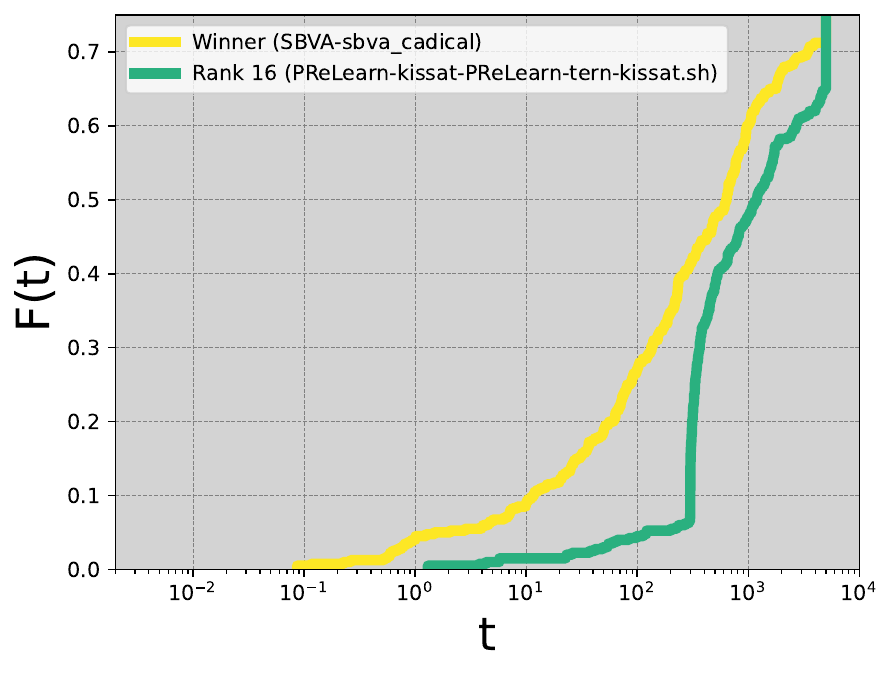}
    \vspace{-.2cm}    
    \caption{The winning SAT solver is FOSD over the 16th ranked ``PReLearn-kissat-PReLearn-tern-kissat.sh''.}
    \label{fig:cdfs_fosd}
\end{figure}

Like all previous work on utilitarian algorithm configuration, we have so far assumed that we know the  utility function $u$ being optimized. In practice, however, we may not be sure how to instantiate certain parameters of $u$. In this section we illustrate a way around such uncertainty: 
gathering runtime data from which we can draw conclusions and then testing how sensitive our conclusions are to the parameter values. 

\begin{figure}
    \centering
    \includegraphics[width=.6\linewidth]{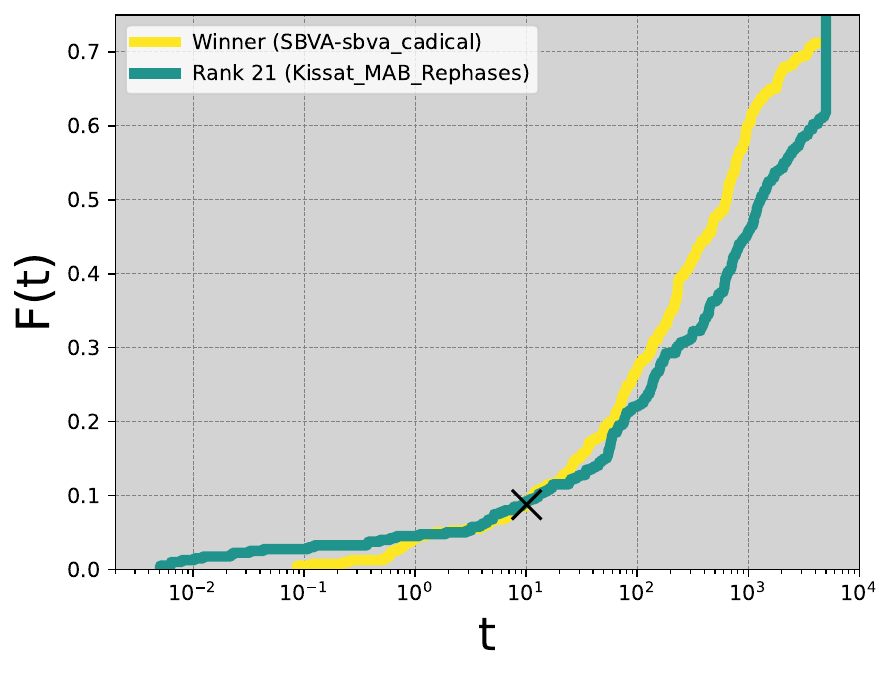}
    \vspace{-.2cm}
    \caption{The winning SAT solver is FOSD over the 21st ranked ``Kissat\_MAB\_Rephases'' for runtimes $> \;\sim 10$s.}
    \label{fig:cdfs_part}
\end{figure}

We illustrate several different approaches to such an analysis by working through a concrete example: the International SAT Competition (\url{https://satcompetition.github.io}). This is a prominent annual event held since at least 2002. Participants in the competition are asked to submit solvers which will be evaluated on an unknown set of SAT problem instances. Typically, 300 to 600 benchmark instances are used, and each solver is given a maximum of 5000 seconds to solve each instance. 
Once the submitted solvers have been run on the benchmark instances, we can derive CDFs for each one. We can infer a great deal from these CDFs about which solver is best under various criteria. 

\begin{figure*}[t]
\includegraphics[height=11em]{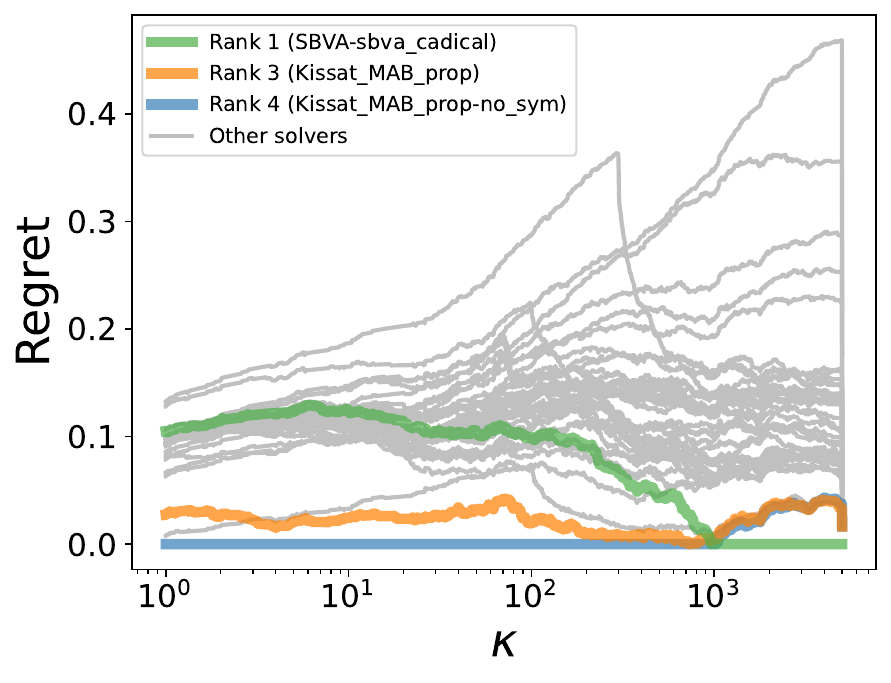}\hfill%
\includegraphics[height=11em]{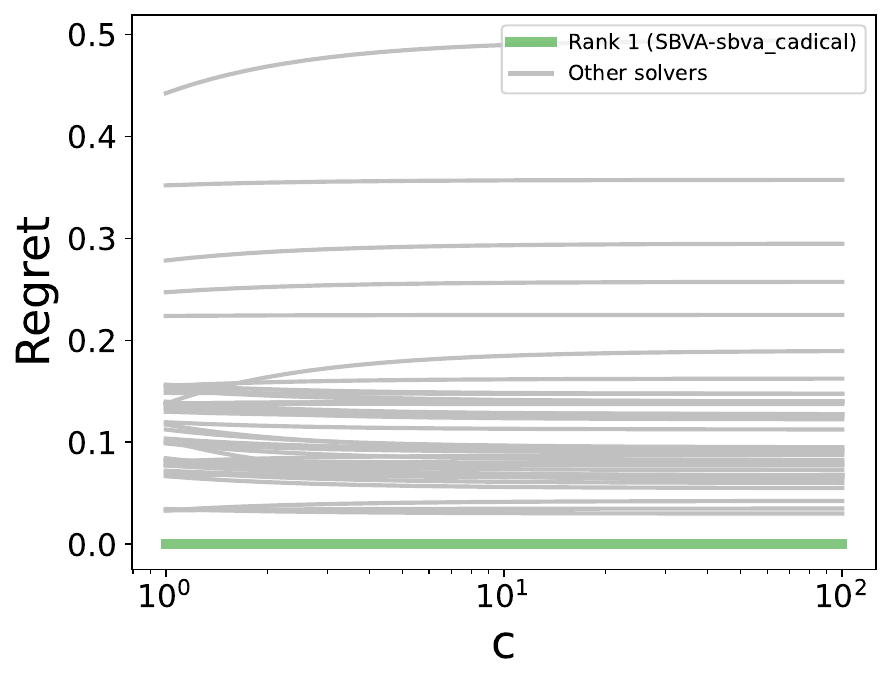}\hfill%
\includegraphics[height=11em]{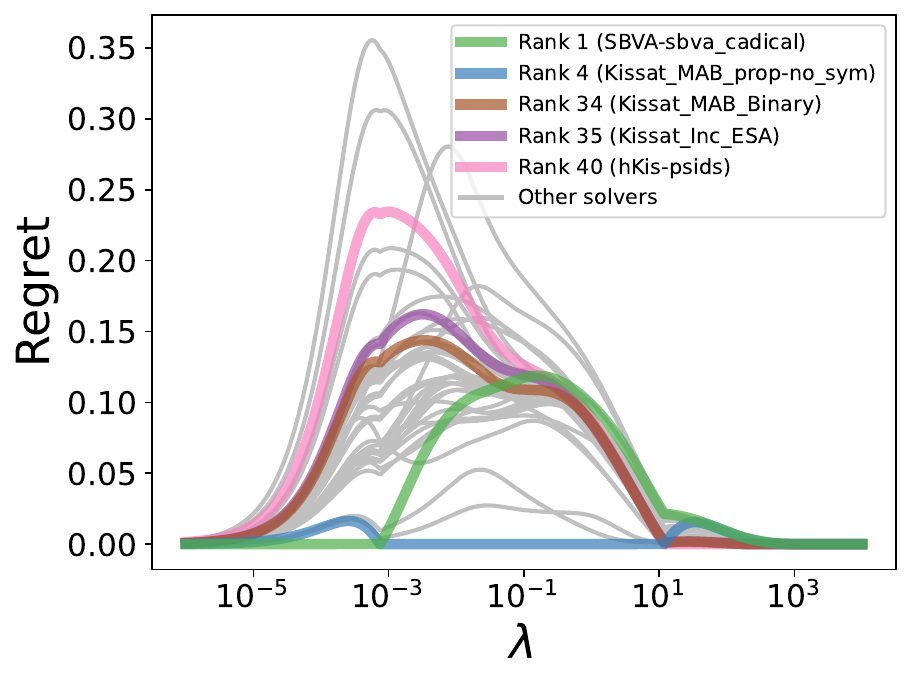}
\vspace{-.3cm}
\caption{Left: Regret using the $PAR(2, \kappa)$ utility function. Center:~Regret using the $PAR(c, 5000)$ utility function. Right:~Regret using the exponential utility function.}
\label{fig:diffc_optimals_utility}
\label{fig:diffk_optimals_utility}
\label{fig:exp_optimals_utility}
\end{figure*}

The simplest inference we can make is based on First Order Stochastic Dominance (FOSD). For two solvers $A$ and $B$ with runtime CDFs $F_A$ and $F_B$, we say that $A$ is FOSD over $B$ if $F_A(t) \ge F_B(t)$ for all $t$ in the support of $A$ and $B$, and $F_A(t) > F_B(t)$ for some $t$. (The inequality is reversed from the definition usually seen because we are dealing with runtime, which we want to minimize, instead of money or some other good that we want to maximize.) An immediate result of this definition is that a solver $A$ is FOSD over a solver $B$ if and only if $A$ is expected-utility preferred over $B$ for \emph{any} utility function. 
For example, consider the runtime data from the 2023 SAT Competition. Only two solvers, the 8th ranked ``PReLearn-kissat-PReLearn-kissat'' and the 16th ranked ``PReLearn-kissat-PReLearn-tern-kissat'' are first-order stochastic dominated by the winner (\cref{fig:cdfs_fosd}). These solvers will never outrank the winner, no matter what evaluation criterion we use. 
Other solvers are ``approximately'' first-order stochastic dominated by the winner. For instance, the 21st ranked solver ``Kissat\_MAB\_Rephases'' is shown in \cref{fig:cdfs_part}, and the winning solver is FOSD over it for runtimes greater than about 10 seconds. (For context, fewer than 10\% of all solves by all solvers took less than 10 seconds.)

If the organizers decide that all runs less than 10 seconds are essentially equivalent, then this will be reflected in the chosen utility function, which will be constant up to (at least) 10 seconds. In this case we are essentially back to having FOSD. The winning solver is expected-utility preferred over solver 21 for any utility function that is constant up to 10 seconds. For instance one of the simplest evaluation criteria we could use in this scenario is the proportion of instances solved. This amounts to optimizing a step function utility of the form $u(t) = 1$ if $t < \kappa$ and $u(t) = 0$ otherwise, where $\kappa$ is a parameter. For any $\kappa > 10$ seconds, the winning solver is FOSD over solver 21. 
In general, a utility function decreases from 1 to 0 as $t$ increases. The regions where the utility function decreases most will be the most important regions. If all of the decrease happens in a region where solver $A$ is FOSD over solver $B$, then $A$ will be expected utility preferred over $B$, regardless of the precise form of the decrease. We expand on this idea in \cref{app:fosd}\shortver{ of the paper's extended version}.

The organizers have often adopted a ``penalized average runtime'' (PAR) scoring function. The score for solving an instance is the time the solver took to solve it, while the score for a timeout is a constant multiplied by the timeout value. This scoring function has two parameters: a penalty factor $c$, and a maximum timeout $\kappa$. We can map this to a utility function in the range $[0, 1]$ by setting $u(t; c, \kappa) = 1 - \frac{t}{c\kappa}$ if $t < \kappa$ and $u(t; c, \kappa) = 0$ otherwise. Ranking solvers according to the PAR($c$, $\kappa$) score is equivalent to (i.e., will always give equivalent rankings as) ranking solvers according to the utility function $u(t;c,\kappa)$. 

In recent years, the SAT Competition has evaluated solvers according to a PAR($c$, $\kappa$) utility function with $c=2$ and $\kappa=5000$. 
We might naturally ask how important the choice of values of these parameters is to the final rankings of solvers, since the values $c=2$ and $\kappa=5000$ are essentially arbitrary. Is the winning solver only the winner because of these specific values? Might the final rankings have been significantly different for different values? If the rankings change significantly with small changes in the parameter values, then we might be less convinced that the winning solver is really the ``best'' one, while if the winning solver remains highly ranked for a wide range of parameter values, we would be more confident about this conclusion.

We can plot the utility of each solver as a function of the parameter(s) of the utility function. We can then look at which solver is best for each value of the parameter(s). \cref{fig:diffk_optimals_utility} (Left) and \cref{fig:diffc_optimals_utility} (Center) show the regret for each solver, which is the difference between its utility and the utility of the best solver for that parameter value. 
Holding $c=2$ fixed, for $\kappa$ above about 1000 the winning solver is best, but for smaller $\kappa$ it is actually the 4th ranked solver (``Kissat\_MAB\_prop-no\_sym'') that is best. For a small range of $\kappa$ around about 1000 the 3rd ranked solver (``Kissat\_MAB\_prop'') is best, but all three solvers have very similar performance at this point. If we are trying to choose a solver for some specific application, but we do not know exactly how much time we will be given to solve an instance, we can choose our solver based on whether we think we will be given more or less than 1000 seconds. However, if we are unsure of whether we have more or less than 1000 seconds, we could choose the 4th ranked solver and our regret would never be too great, whereas if we chose the winning solver our regret could be relatively large. Finally, we can see that when $\kappa=5000$, the best ranked solver is best for all values of $c$. This sort of reasoning can be helpful when we have uncertainty about our utility function's parameters. 

We can also consider utility function other than $PAR(c, \kappa)$. For example, we could use an exponential function: $u(t) = e^{-\lambda t}$, where $\lambda$ is a parameter. If we know the exact value of $\lambda$ that we want to use, then we can just compute the mean utility for each solver and find which one is best. Otherwise we can plot the mean utility of each solver as a function of $\lambda$ and see how the ranking changes, as in \cref{fig:exp_optimals_utility} (Right). For a wide range of parameter values the top-ranked solver (``SBVA-sbva\_cadical'') and the 4th-ranked solver (``Kissat\_MAB\_prop-no\_sym'') are best according to the exponential utility function as well.

\section{Conclusion}

We have proposed four improvements to the utilitarian algorithm configuration procedure COUP that boost its empirical performance without degrading its theoretical bounds. We have shown that these improvements make a real difference to the tightness of the bounds proved by COUP, as well as to the quality of the returned configuration. We demonstrated that COUP is competitive with SMAC, the most widely used heuristic configuration procedure, while also offering theoretical optimality guarantees that improve with time. We are now inclined to believe that utilitarian algorithm configuration methods that offer theoretical guarantees have reached the point that they should be preferred for use in practice. We also described a case study on the SAT Competition that shows how to reason about the robustness of a conclusion recommended by a utilitarian configuration procedure with respect to different utility functions that could have been chosen, implying that uncertainty about the exact utility function used need not be a barrier to the adoption of utilitarian algorithm configuration approaches.  In future work we plan to perform a more extensive experimental evaluation to make this claim definitive, comparing COUP's performance with SMAC across a wider set of utility functions, and also contrasting COUP with other heuristic methods.

\clearpage
\section*{Acknowledgments}
We would like to thank Greg d'Eon for helpful discussions and suggestions regarding plots.

\bibliography{aaai2026}

\clearpage
\appendix
\section{Algorithm Pseudocode}\label{app:code}

Pseudocode for the original COUP, as reported in \citet{graham2024utilitarian}, is presented in \cref{alg:coup}. Pseudocode for our improved version of COUP is presented in \cref{alg:coupplus}. We discuss particular differences between the two with respect to the four improvements we have suggested. Define $d(p, q) = p \log(p / q) + (1 - p) \log ((1 - p) / (1 - q))$.

\subsection{Improved Confidence Bounds}

\paragraph{COUP:} The term $\alpha_p$ (defined in \citet{graham2024utilitarian}) is explicitly chosen to satisfy Hoeffding's inequality in the bounds defined on lines 26 and 27 (and 12 and 13). 

\paragraph{COUP+:} The minimization and maximization terms defining the bounds on lines 19-21 appeal directly to the KL term $d$. When solved numerically, these bounds are always at least as good as their Hoeffding equivalents. 

\subsection{Lower Upper Confidence Bound}

\paragraph{COUP:} 
In each iteration, the configuration with largest UCB is chosen at line 16.

\paragraph{COUP+:} 
In each iteration, the configuration with largest empirical mean, and the remaining configuration with largest UCB are chosen at lines 7 and 8.  

\subsection{Adding New Configurations}

\paragraph{COUP:} 
COUP takes a sequence of pairs parameters $(\epsilon_p, \gamma_p)$ for $p = 1, 2, 3, ...$, which determine how many new configurations to add (line 5) and when to add them (line 15).

\paragraph{COUP+:} 
We do away with the need to provide this sequence of parameters (an onerous challenge) and instead add a single new configuration when a natural condition is satisfied (line 28). 

\subsection{Model-guided Configuration Search}

\paragraph{COUP:} 
No model is used in basic COUP. 

\paragraph{COUP+:} 
The model is trained on line 30 and a configuration is sampled from it on line 31.

\begin{algorithm*}[t]
\caption{COUP}\label{alg:coup}
\begin{algorithmic}[1]    
    \State \textbf{Input:} distribution over configurations $\D_\A$; instances $j = 1, 2, ...$; utility function $u$; failure parameter $\delta$; phase parameters $\{\epsilon_p, \gamma_p\}_{p=1,2,3,...}$.
    \State $n_0 \gets 0$
    \State $\A_0 \gets \emptyset$
    \For{$p = 1, 2, 3, ...$ until terminated}
        \State $n_p \gets \bigg\lceil \frac{\ln\frac{\pi^2 p^2}{3 \delta}}{\gamma_p} \bigg\rceil$
        \State $N_p \gets$ sample $n_p - |\A_{p-1}|$ new configurations from $\D_\A$ 
        \State $\A_p \gets \A_{p-1} \cup N_p$ \Comment{all configurations for this phase}
        \For{$i \in N_p$} \Comment{initializations for new configurations}
            \State $LCB_i \gets 0$, $UCB_i \gets 1$; $\widehat{U}_i \gets 0$; $\widehat{F}_i \gets 0$; $m_i \gets 0$; $\kappa_i \gets 1$
        \EndFor
        \For{$i \in \A_{p-1}$} \Comment{update bounds for existing configurations}            
            \State $UCB_{i} \gets \widehat{U}_{i} + \big(1-u(\kappa_i)\big) \cdot \alpha_p(m_i, \kappa_i)$
            \State $LCB_{i} \gets \widehat{U}_{i} - \alpha_p(m_i, \kappa_i) - u(\kappa_i)(1 - \widehat{F}_i)$
        \EndFor
        \While{$\max_{i \in \A_p} UCB_i - \max_{i \in \A_p} LCB_i \ge \epsilon_p$} \Comment{phase termination condition}
            \State $i \gets \arg\max_{i' \in \A_p} UCB_{i'}$\label{line:imany}
            \State $m_i \gets m_i + 1$
        
            \If{$2 \alpha_p(m_i, \kappa_i) \le u(\kappa_i)(1 - \widehat{F}_i)$} \Comment{captime doubling condition}
                \State $\kappa_i \gets 2 \kappa_i$
                \State $t_{i1}(\kappa_i), ..., t_{im_i}(\kappa_i) \gets $ runtime of $i$ on instances $1, ..., m_i$ with timeout $\kappa_i$
            \Else
                \State $t_{im_i}(\kappa_i) \gets $ runtime of $i$ on instance $m_i$ with timeout $\kappa_i$
            \EndIf
            \State $\widehat{F}_{i} \gets \frac{|\{ j \in [m_i] \;:\; t_{ij}(\kappa_i) < \kappa_i \}|}{m_i}$ \Comment{fraction of runs that completed}
            \State $\widehat{U}_{i} \gets \frac{1}{m_i} \sum_{j=1}^{m_i} u\big( t_{ij}(\kappa_i) \big)$ \Comment{empirical average utility}
            \State $UCB_{i} \gets \widehat{U}_{i} + \big(1-u(\kappa_i)\big)\cdot\alpha_p(m_i, \kappa_i)$
            \State $LCB_{i} \gets \widehat{U}_{i} - \alpha_p(m_i, \kappa_i) - u(\kappa_i)(1 - \widehat{F}_i)$
        \EndWhile
    \EndFor    
    \State \textbf{return} $\arg\max_{i\in \A_p} LCB_i$
\end{algorithmic}
\end{algorithm*}

\begin{algorithm*}[t]
\caption{COUP+}\label{alg:coupplus}
\begin{algorithmic}[1]
    \State \textbf{Input:} distribution over configurations $\D_\A$; instances $j = 1, 2, ...$; utility function $u$; failure parameter $\delta$; initial set of configurations $n_0$.
    \State $\A \gets$ sample $n_0$ configurations at random
    \For{$i \in A$} \Comment{initializations}
        \State $\widehat{U}_i \gets 0$, $\widehat{U}_i^{(ucb)} \gets 1$, $\widehat{U}_i^{(lcb)} \gets 0$, $\widehat{F}_i \gets 0$, $\widehat{F}_i^{(lcb)} \gets 0$, $UCB_i \gets 1$, $LCB_i \gets 0$, $m_i \gets 0$, $\kappa_i \gets 1$
    \EndFor
    \While{not interrupted}
        \State $i_1 \gets \arg\max_{i' \in \A} \widehat{U}_{i'}$ 
        \State $i_2 \gets \arg\max_{i' \in \A \setminus \{i_1\}} UCB_{i'}$ 
        \For{$i \in \{i_1, i_2\}$}
            \If{$\widehat{U}_i^{(ucb)} - \widehat{U}_i^{(lcb)} \le u(\kappa_i)\big( 1 - \widehat{F}_i^{(lcb)} \big)$} \Comment{captime doubling condition}
                \State $\kappa_i \gets 2 \kappa_i$
                \State $t_{i,1}(\kappa_i), ..., t_{i,m_i-1}(\kappa_i) \gets $ runtime of $i$ on instances $1, ..., m_i-1$ with timeout $\kappa_i$
            \EndIf
            \State $t_{i,m_i}(\kappa_i) \gets $ runtime of $i$ on instance $m_i$ with timeout $\kappa_i$
            \State $m_i \gets m_i + 1$
            \State $\widehat{U}_i \gets \frac{1}{m_i} \sum_{j=1}^{m_i} u\big(t_{i, j}(\kappa_i)\big)$ \Comment{capped sample average utility}
            \State $\widehat{F}_i \gets \frac{|\{ j \;:\; t_{i, j}(\kappa_i) < \kappa_i \}|}{m_i} $ \Comment{fraction of completed runs}
            \State $a \gets \frac{1}{m_i}\ln\frac{36 n^2 m_i^2 (\log(\kappa_i + 1)^2}{\delta}$
            \State $\widehat{U}_i^{(ucb)} \gets \max\{ u \in [0, 1] \;:\; d(\widehat{U}_i, u) \le a\}$
            \State $\widehat{U}_i^{(lcb)} \gets \min\{ u \in [0, 1] \;:\; d(\widehat{U}_i, u) \le a\}$
            \State $\widehat{F}_i^{(lcb)} \gets \min\{ u \in [0, 1] \;:\; d(\widehat{F}_i, u) \le a\}$
            \State $UCB_i \gets \widehat{U}_i^{(ucb)}$
            \State $LCB_i \gets \widehat{U}_i^{(lcb)} - u(\kappa_i)\big(1 - \widehat{F}_i^{(lcb)}\big)$
        \EndFor
        \State $\epsilon \gets \max_{i \in \A} UCB_{i} - \max_{i \in \A} LCB_{i}$
        \State $n' \gets n_0 + \lfloor \frac{n - n_0}{2} \rfloor$ \Comment{number of randomly-selected configurations}
        \State $\gamma \gets \frac{1}{n'}\ln \frac{\pi^2 n'^2 }{3\delta} $
        \If{$ \epsilon^2 < \gamma \big( 1 - \max_{i \in \A}UCB_{i} \big)$} \Comment{time to add next configuration}
            \If{$n$ is even } 
                \State $\mathcal{M} \gets$ train model on observed runtime data
                \State $i \gets $ sample configuration from $\mathcal{M}$
            \Else
                \State $i \gets $ sample a random configuration
            \EndIf
            \State $\widehat{U}_i \gets 0$, $\widehat{U}_i^{(ucb)} \gets 1$, $\widehat{U}_i^{(lcb)} \gets 0$, $\widehat{F}_i \gets 0$, $\widehat{F}_i^{(lcb)} \gets 0$, $UCB_i \gets 1$, $LCB_i \gets 0$, $m_i \gets 0$, $\kappa_i \gets 1$
            \State $\A \gets \A \cup \{i\}$ 
            \State $n \gets n + 1$
        \EndIf
    \EndWhile    
    \State \textbf{return} $\arg\max_{i\in \A} LCB_i$
\end{algorithmic}
\end{algorithm*}

\section{Deriving Confidence Bounds}\label{app:confbounds}

Suppose we have a lower confidence bounds on the CDF $F$ at $\kappa$:
\begin{align*}
    F^{(lcb)} \le F(\kappa).
\end{align*}
It doesn't matter where this bound comes from; it may be from Hoeffding's inequality, or from the KL-term directly, or from somewhere else. Simple arithmetic then implies that 
\begin{align*}
    1 - F^{(lcb)} \ge 1 - F(\kappa)
\end{align*}
and so 
\begin{align*}
    - u(\kappa)\big(1 - F^{(lcb)}\big) \le - u(\kappa)\big(1 - F(\kappa)\big) .
\end{align*}
Meanwhile, the law of total expectation says that
\begin{align*}
    \E[u(\min(t, \kappa))] &= \E[u(\min(t, \kappa)) | t \le \kappa]F(\kappa) \\
    &\quad + \E[u(\min(t, \kappa)) | t > \kappa](1 - F(\kappa)) \\
    &= \E[u(t) | t \le \kappa]F(\kappa) + u(\kappa)(1 - F(\kappa)) \\
    &= \E[u(t) | t \le \kappa]F(\kappa) + u(\kappa)(1 - F(\kappa)) \\
    &\quad + \E[u(t) | t > \kappa](1 - F(\kappa)) \\
    &\quad - \E[u(t) | t > \kappa](1 - F(\kappa)) \\
    &= \E[u(t)] + u(\kappa)(1 - F(\kappa)) - \\
    &\quad \E[u(t) | t > \kappa](1 - F(\kappa))
\end{align*}
And so rearranging gives 
\begin{align*}
    \E[u(t)] &= \E[u(\min(t, \kappa))] - u(\kappa)(1 - F(\kappa)) \\
    &\quad + \E[u(t) | t > \kappa](1 - F(\kappa)) .
\end{align*}
Using this, and the above bound on $- u(\kappa)\big(1 - F(\kappa)\big)$, and the fact that $\E[u(t) | t > \kappa](1 - F(\kappa))$ is positive, gives:
\begin{align*}
    \E[u(t)] &= \E[u(\min(t, \kappa))] - u(\kappa)(1 - F(\kappa)) \\
    &\quad + \E[u(t) | t > \kappa](1 - F(\kappa)) \\
    &\ge \E[u(\min(t, \kappa))] - u(\kappa)(1 - F(\kappa)) \\
    &\ge \E[u(\min(t, \kappa))] - u(\kappa)\big(1 - F^{(lcb)}\big) .
\end{align*}

\section{Incorporating Configurations From the Model}\label{app:gamma}

We assume we have a set of configurations $\A$, possibly uncountably infinite, and a distribution $\D_\A$ over $\A$ that we can sample configurations from. For any $\gamma \in (0, 1)$ let $OPT^{\gamma} = \sup\big\{ \mu \;:\; \Pr_{a \sim \D_\A}[U_a \le \mu] \le 1 - \gamma \big\}$, where $U_a$ is the mean utility of configuration $a$. Then $OPT^\gamma$ is the utility of the best configuration that remains after we have excluded the top $\gamma$-fraction of configurations. If we sample $n$ configurations at random, then we can derive a guarantee that we are unlikely to sample a configuration that is much better. Set $\gamma = \frac{1}{n} \log{\frac{\pi^2n^2}{3 \delta}}$. For each configuration $i = 1, ..., n$, the probability that $U_i \le OPT^{\gamma}$ is at most $1 - \gamma$. So the probability that all $n$ configurations have $U_i \le OPT^{\gamma}$ is at most $(1 - \gamma)^n$. Using the inequality $1 - x \le e^{-x}$, the probability that all $n$ configurations have $U_i \le OPT^{\gamma}$ is at most $e^{-n\gamma} \le \frac{3 \delta}{\pi^2n^2}$. A union bound then says that at all points during COUP's execution, the probability that all sampled configurations have $U_i \le OPT^{\gamma}$ is at most $\frac{\delta}{2}$.

If we use the model to recommend half of the added configurations, then we in fact only have a sample of $n' = n_0 + \lfloor \frac{n - n_0}{2} \rfloor \ge \frac{n}{2}$ random configurations, where $n_0$ is the number of initial configurations considered, before the model is trained. If we then set $\gamma' = \frac{1}{n'} \log{\frac{\pi^2n'^2}{3 \delta}}$ we can see that our guaranteed $\gamma$ degrades by just a constant factor, but will still be driven to 0.

\section{Comparison with Other Theoretically-motivated Procedures}\label{app:pgap}

\begin{figure*}
\includegraphics[width=0.34\textwidth]{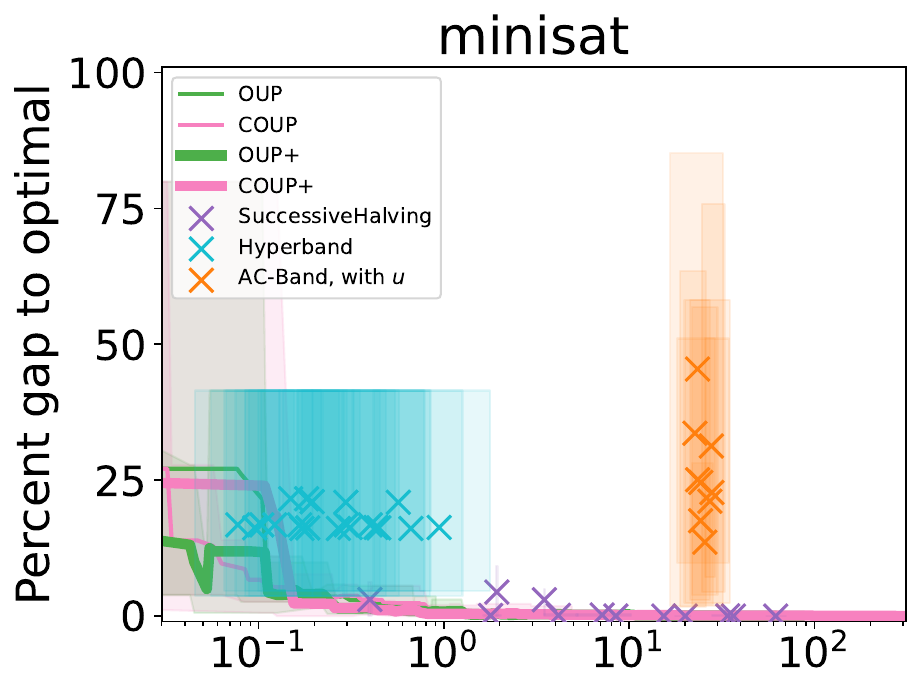}\hfill%
\includegraphics[width=0.32\textwidth]{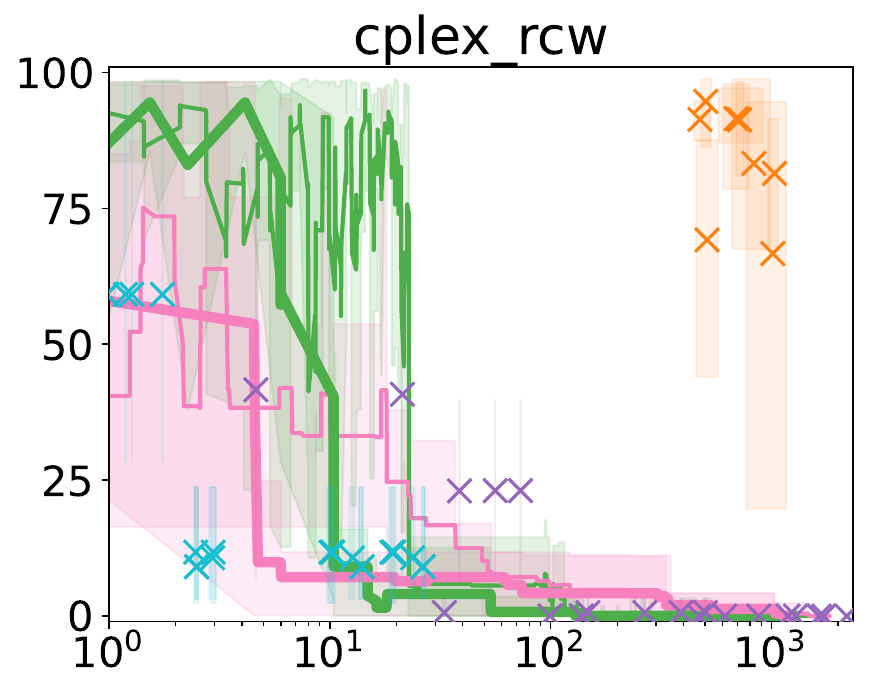}\hfill%
\includegraphics[width=0.32\textwidth]{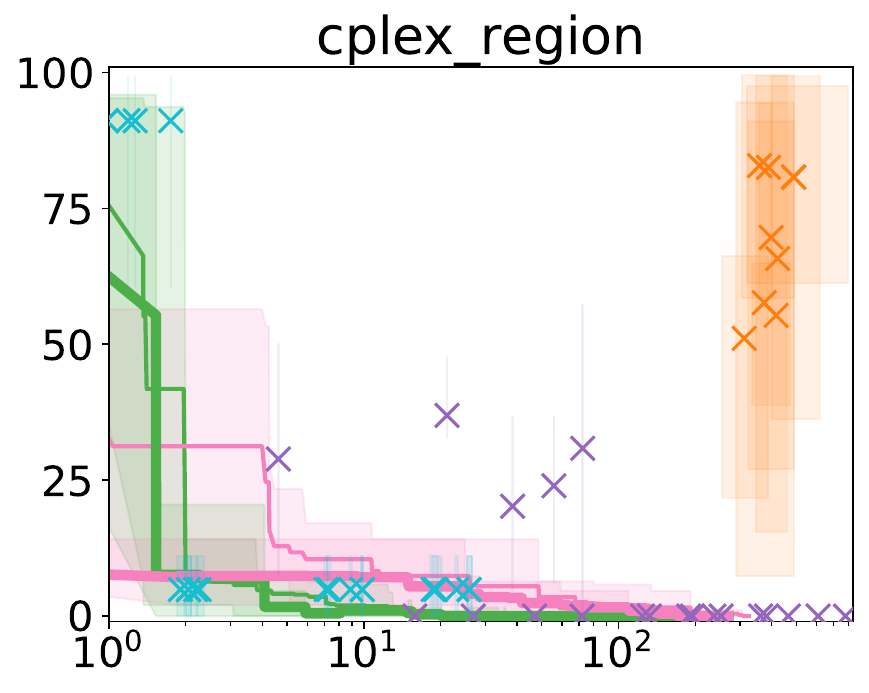}\\
\includegraphics[width=0.34\textwidth]{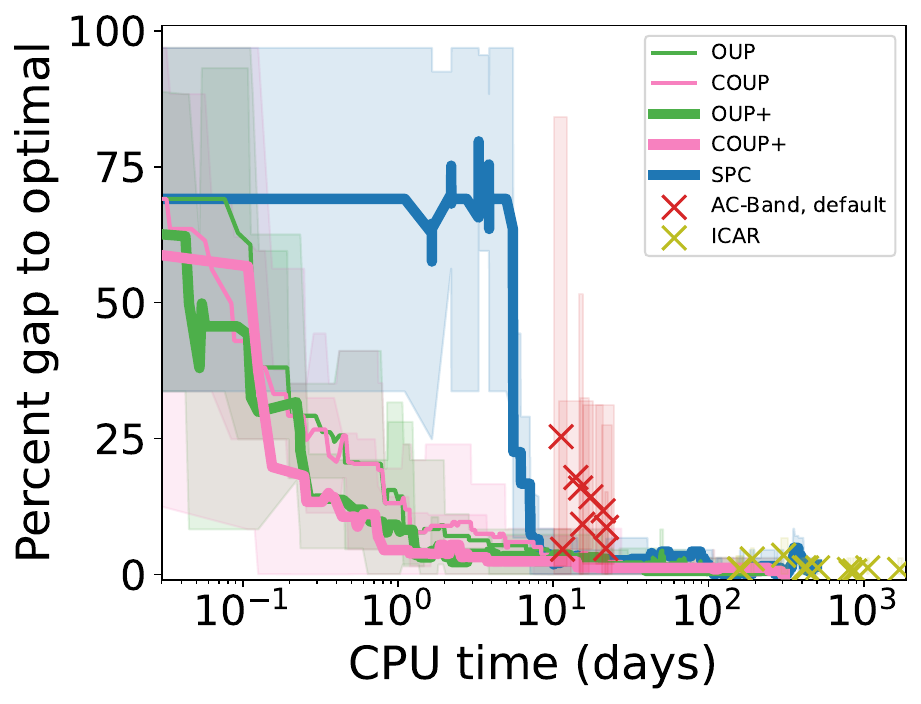}\hfill%
\includegraphics[width=0.32\textwidth]{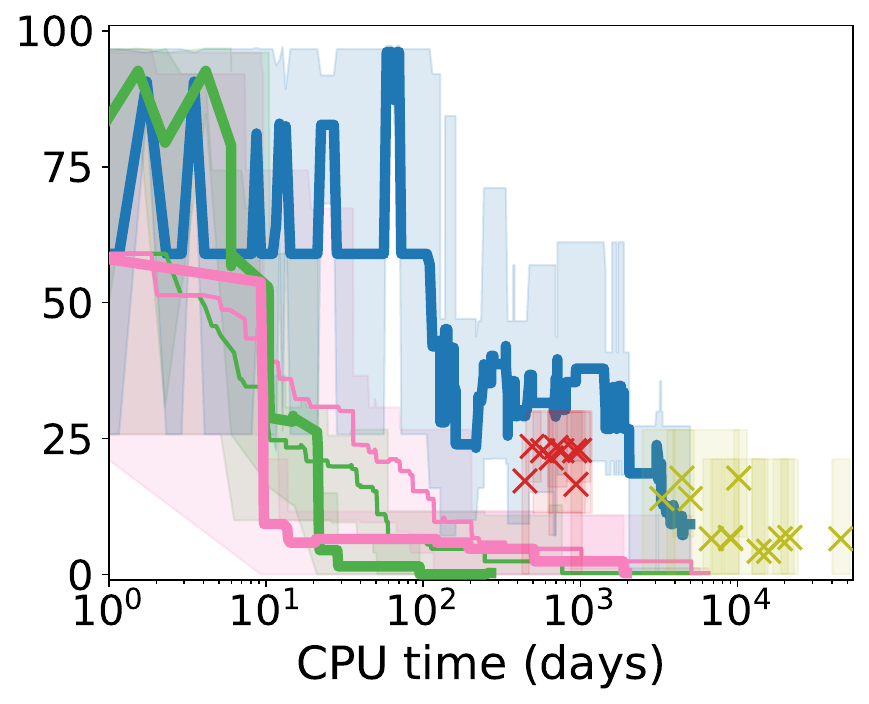}\hfill%
\includegraphics[width=0.32\textwidth]{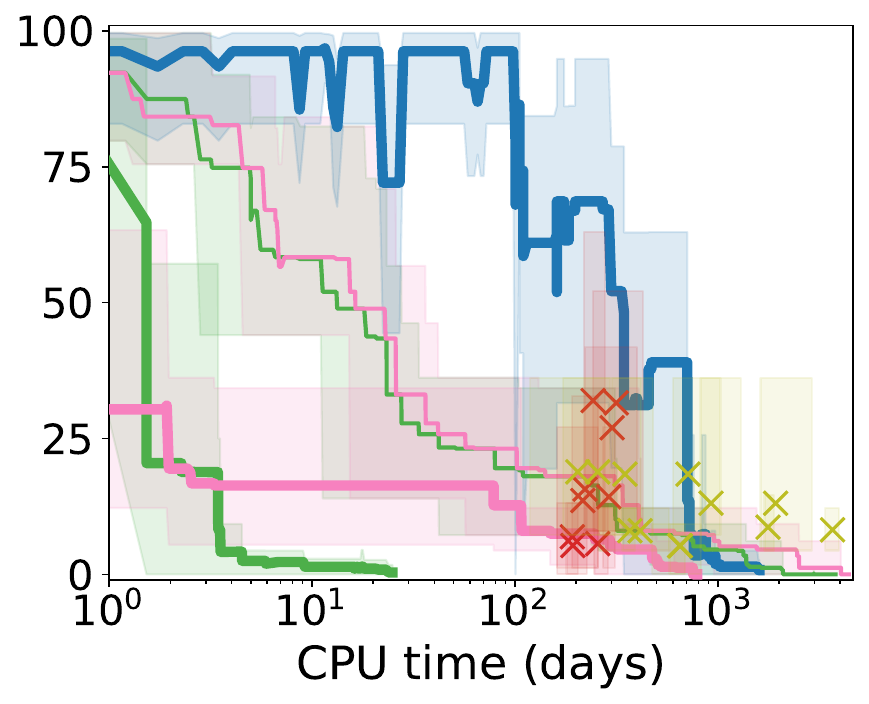}
\caption{Figure 5 reproduced from \citet{graham2024utilitarian} with our refinements added to OUP and COUP. Top row: Procedures that can optimize any utility function (log-Laplace utility function was used). Bottom Row: Procedures that optimize some kind of capped average runtime.}
\label{fig:pgap_plots}
\end{figure*}

\cref{fig:pgap_plots} shows a comparison of our improved variants COUP and OUP against a number of other theoretically-motivated configuration procedures. This is essentially a reproduction of Figure 5 from \citet{graham2024utilitarian} with out new COUP and OUP variants included.

\section{Details of the Prediction Model}\label{app:model}

We use the XGBoost \citep{chen2016xgboost} for implementing the prediction model. This allows us to predict a configuration's mean utility. We produce uncertainty estimates by performing a bootstrap aggregation procedure. We subsample 100 sets of configurations of size twice the number of configurations, and train an XGBoost model on each one. For each candidate configuration we make a prediction with each of the 100 models, giving us a distribution of values. We then consider 95\% confidence intervals, giving us an uncertainty band around each prediction. We do 100 rounds of boosting, giving 100 trees, which we limit to a depth of 3. We train the model with a squared error loss. 

\section{Deriving Utility Functions from Observed Runtime Data}\label{app:functions}

A user will sometimes find themselves in a position where they know the general form of the utility function they would like to optimize, but not exactly know certain parameters of it. For example, we may know that we will only have a fixed amount of time to run our algorithm, but not be certain about exactly how much time this will be. Will we have 10 hours, or 12 hours? Or only 5 hours? Rather than force the user to specify the value of this unknown parameter, we are interested in making statements about which algorithm is best \emph{given the value of this parameter}. Perhaps the same algorithm is best as long as we are given between 6 and 13 hours to run it. In this case, specifying the exact value of the parameter is not necessary. 

As a concrete example, consider the International SAT Competition\footnote{https://satcompetition.github.io/}. This is an annual event held since at least 2002. Participants in the competition are asked to submit solvers which will be evaluated on an unknown set of SAT problem instances. Typically, 300 to 600 benchmark instances are used, and each solver is given a maximum of 5000 seconds to solve each instance. 




In many years, the organizers have adopted a ``penalized average runtime'' (PAR) scoring function. The score for solving an instance is the time the solver took to solve it, while the score for a timeout is a constant times the timeout value. This scoring function has two parameters: a penalty factor $c$, and a maximum timeout $\kappa$. The score of a run that takes $t$ seconds is $s(t) = t$ if $t < \kappa$ and $s(t) = c\kappa$ otherwise.

For consistency with other scoring metrics we might be interested in, we can map this to a \emph{utility function} in the range $[0, 1]$ as follows.
\begin{align*}
    u(t; c, \kappa) = \begin{cases}
        1 - \frac{t}{c\kappa} &\quad \text{ if } t < \kappa \\
        0 &\quad \text{ otherwise }
    \end{cases}
\end{align*}
Minimizing the PAR($c$, $\kappa$) score is equivalent to (i.e., will always give equivalent rankings as) maximizing the utility function $u(t;c,\kappa)$. In recent years, the SAT Competition has evaluated solvers according to a PAR($c$, $\kappa$) utility function with $c=2$ and $\kappa=5000$. 

We might naturally ask how important the choice of values of these parameters is to the final rankings of solvers. Is the winning solver only the winner because of these specific values? Might the final rankings have been significantly different for different values? If we had used a penalty factor of $c=3$, or a timeout of $\kappa=4000$, would the picture have changed? It is important to ask this question, since the values $c=2$ and $\kappa=5000$ are essentially arbitrary. If the rankings change significantly with small changes in the parameter values, then we are somehow less convinced that the winning solver is really the ``best'' one. On the other hand, if the winning solver remains highly ranked for a wide range of parameter values, then we are more confident in our conclusion that it is indeed the best solver.

\begin{figure}
    \centering
    \includegraphics[width=.9\linewidth]{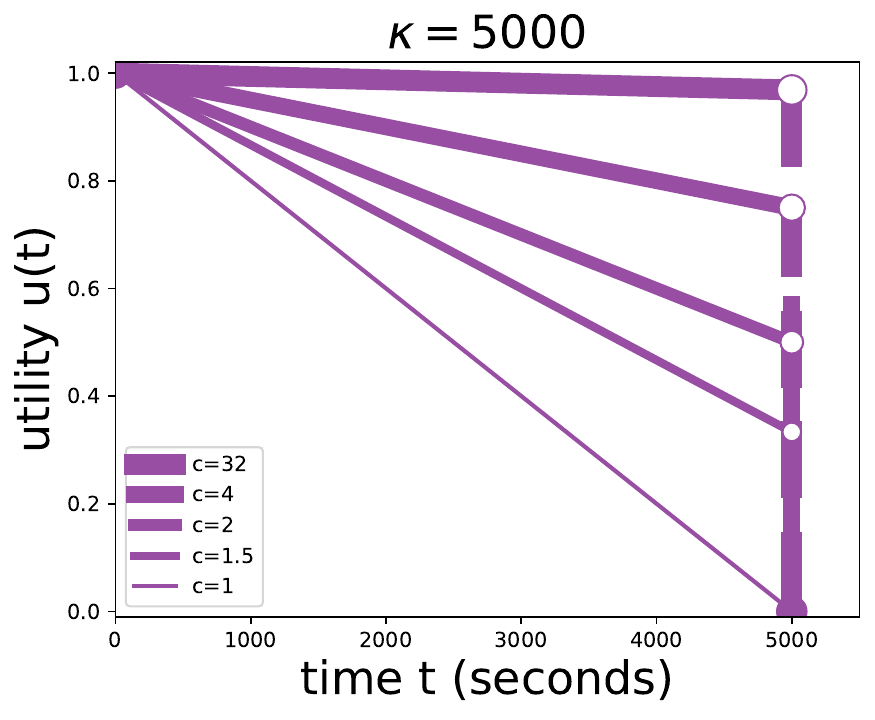}
    \hfill 
    \includegraphics[width=.9\linewidth]{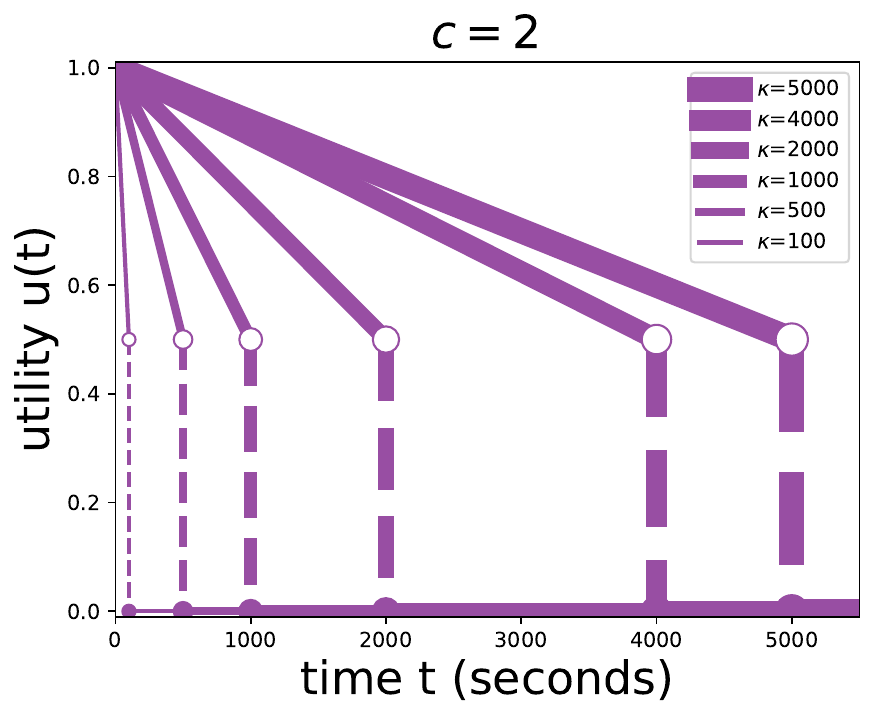}
    \caption{\small PAR($c$, $\kappa$) utility functions with different parameters.}
    \label{fig:udiff}
\end{figure}

\begin{figure}
    \centering
    \includegraphics[width=.9\linewidth]{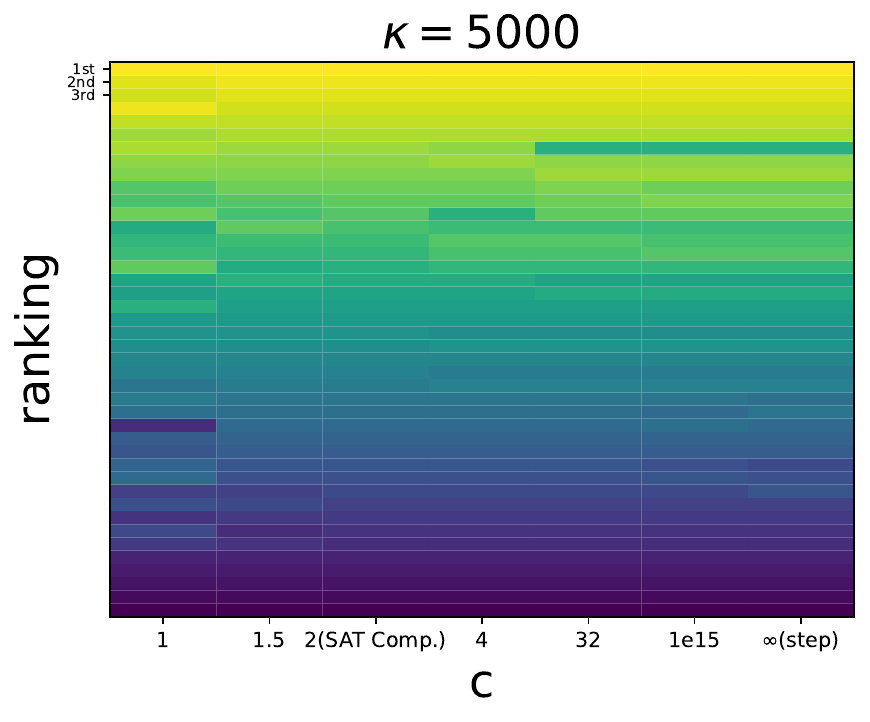}
    \hfill 
    \includegraphics[width=.9\linewidth]{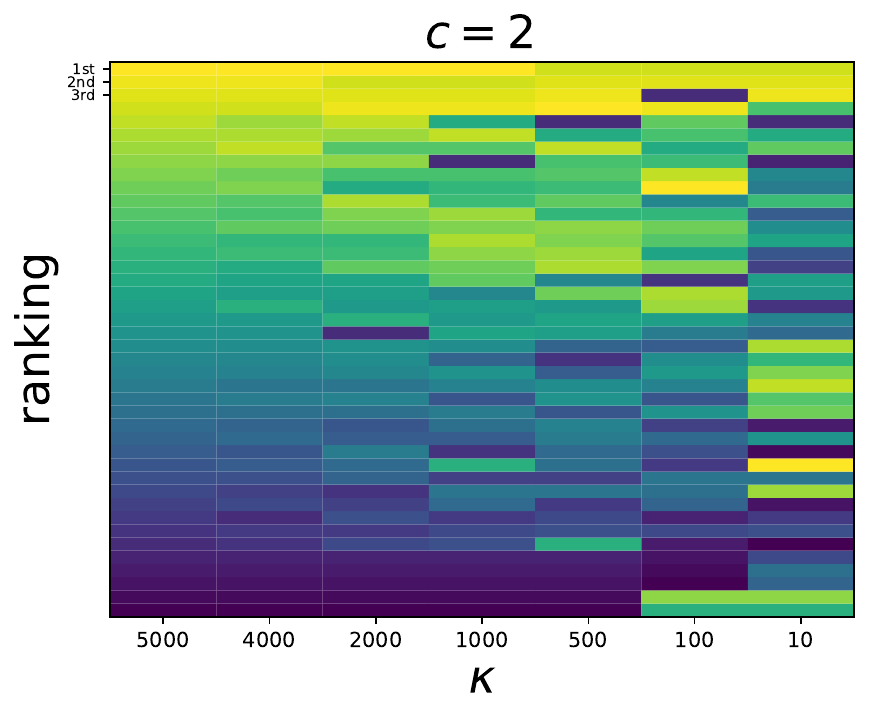}
    \caption{\small SAT solvers ranked by PAR($c$, $\kappa$) utility functions with different parameters.}
    \label{fig:rankdiff}
\end{figure}

In general, as we vary $c$ and $\kappa$, we get a family of qualitatively different utility functions, as shown in \cref{fig:udiff}. How much do the rankings change with different $c$ and $\kappa$ values? We can assign each solver a color, which we choose based on its 2023 SAT Competition PAR-2 ranking. Then we can easily see how different parameter values change the final ranking of solvers. \cref{fig:rankdiff} shows these different rankings. 

Visually, the rankings for different $c$ look quite similar, while those for different $\kappa$ can look significantly different. We can actually measure the difference between two rankings using the L1- or ``Manhattan'' distance. If $x = (x_1, ..., x_n)$ is a ranking of $n$ solvers, so that $x_i$ is the rank given to solver $i$, and $y = (y_1, ..., y_n)$ is another ranking, then the L1-distance between $x$ and $y$ is the sum of the absolute differences between the rankings given to each solver. That is $d(x, y) = \big| x_1 - y_1 \big| + ... + \big | x_n - y_n \big|$. (This is also sometimes known as the ``Spearman's Footrule" distance for rankings.) We can look at the L1-distances between the rankings given by different values of $c$, shown in \cref{fig:l1diff}.

\begin{figure}
    \centering
    \small
    \tabcolsep=0.15cm
    \begin{tabular}{ |c|ccccccc| }
        \hline
        c & 1 & 1.5 & 2 & 4 & 32 & 1e15 & $\infty$ \\
        \hline
        1 & 0 & 48 & 56 & 72 & 76 & 82 & 84 \\
        1.5 & $\cdot$ &  0 & 10 & 26 & 34 & 38 & 40 \\
        2 & $\cdot$ & $\cdot$ &  0 & 16 & 26 & 30 & 32 \\
        4 & $\cdot$ & $\cdot$ & $\cdot$ &  0 & 12 & 18 & 20 \\
        32 & $\cdot$ & $\cdot$ & $\cdot$ & $\cdot$ &  0 &  8 & 10 \\
        1e15 & $\cdot$ & $\cdot$ & $\cdot$ & $\cdot$ & $\cdot$ &  0 &  8 \\
        $\infty$ & $\cdot$ & $\cdot$ & $\cdot$ & $\cdot$ & $\cdot$ & $\cdot$ &  0 \\
        \hline
    \end{tabular}
    \begin{tabular}{ |c|ccccccc| } 
        \hline
        $\kappa$ & 5000 & 4000 & 2000 & 1000 & 500 & 100 & 10 \\
        \hline
        5000 &  0 &  30 &  94 & 170 & 204 & 288 & 504 \\
        4000 &  $\cdot$ &   0 &  80 & 162 & 198 & 286 & 504 \\
        2000 &  $\cdot$ &  $\cdot$ &   0 & 110 & 154 & 252 & 466 \\
        1000 & $\cdot$ & $\cdot$ & $\cdot$ &   0 &  78 & 198 & 410 \\
         500 & $\cdot$ & $\cdot$ & $\cdot$ &  $\cdot$ &   0 & 154 & 366 \\
         100 & $\cdot$ & $\cdot$ & $\cdot$ & $\cdot$ & $\cdot$ &   0 & 270 \\
          10 & $\cdot$ & $\cdot$ & $\cdot$ & $\cdot$ & $\cdot$ & $\cdot$ &   0 \\
        \hline
    \end{tabular}
    \caption{\small L1-distances between $PAR(c, \kappa)$ rankings for different values of $c$ and $\kappa$.}
    \label{fig:l1diff}
\end{figure}

With $n = 42$ solvers, the maximum distance possible is $\frac{n(n-1)}{2} = 861$, which gives us a sense of how large these distances are. The first thing we observe is that the rankings for different $c$ are relatively close, while the rankings for different $\kappa$ can be relatively far. This reinforces what we see in \cref{fig:rankdiff}. We can see that actually the SAT Competition scoring function is not too far (distance 32) from a simple step function at $\kappa = 5000$. For timeouts less than about $\kappa = 1000$ the ratings can be very different. Of course in general we can change $c$ and $\kappa$ simultaneously. We can be more concrete about which solver is best given the values of these parameters.

\subsection{Utility functions and First-order Stochastic Dominance}\label{app:fosd}

Once the submitted SAT solvers have been run on the set of input instances, the organizers have data on the runtime required by each solver. This data gives us a cumulative distribution function (CDF) which describes the distribution of solve times for each solver. Using these CDFs, we can infer a good deal about which solver is best by a variety of metrics.

We have two algorithms $A$ and $B$ and we want to choose the ``fastest'' one. If each of them always takes the same amount of time, then we just choose the one whose time is shortest, but if they take a variable amount of time, say because they are randomized, or because they will be run on future inputs of unknown difficulty, then it is not so straightforward. In this case, each algorithm will be associated with a runtime distribution.

Given two runtime distributions with support in the non-negative reals and CDFs $F_A$ and $F_B$, we say that $A$ is \emph{first-order stochastic dominant} (FOSD) over $B$ if $F_A(t) \ge F_B(t)$ for all $t$ in the support of $A$ and $B$, and $F_A(t) > F_B(t)$ for some $t$.\footnote{Note that the inequality is reversed from the definition usually seen because we are dealing with runtime, which we want to minimize, instead of money or some other good that we want to maximize.} An immediate result of this definition is that a solver $A$ is FOSD over a solver $B$ if and only if $A$ is expected-utility preferred over $B$ for any utility function. This is the first and simplest rule we can apply to the CDFs of two solvers to infer a broad class of utility functions for which $A$ is preferred:
\begin{quote}
    \emph{If $A$ is FOSD over $B$, then $A$ is preferred to $B$ for any utility function.}    
\end{quote}
Only two solvers, the 8th ranked ``PReLearn-kissat-PReLearn-kissat'' and the 16th ranked ``PReLearn-kissat-PReLearn-tern-kissat'') are first-order stochastic dominated by the winner (\cref{fig:cdfs_fosd}). 

Other solvers are ``approximately'' first-order stochastic dominated by the winner. The 21st ranked solver (``Kissat\_MAB\_Rephases'') is shown in \cref{fig:cdfs_part}, and the winning solver is FOSD over it for runtimes greater than about 10 seconds. For context, fewer than 10\% of all solves by all solvers took less than 10 seconds. Meanwhile, The winning solver is FOSD over \emph{all} solvers for 

As we will see, if we are in a position where runs less than 10 seconds are irrelevant to us, then this will be reflected in our utility function, which will be constant up to (at least) 10 seconds, and we are essentially back in the first case where we have FOSD. The winning solver is expected-utility preferred over solver 21 for any utility function that is constant up to 10 seconds.

For simplicity, let's start by ranking the solvers according to different simple step utility functions: $u(t;\kappa) = 1$ if $t < \kappa$ and $u(t;\kappa) = 0$ otherwise. A general, cartoon version of the scenario with the ``approximately'' FOSD CDFs in \cref{fig:cdfs_part} is shown in \cref{fig:cdf_cartoon}. The expected utility of any solver $s$ using a step function at $\kappa$ is 
\begin{align*}
    U_s(\kappa) &\defeq \int_{t=0}^\infty u(t ; \kappa) dF_s(t) \\
    &= \int_{t=0}^\kappa dF_s(t) \\
    &= F_s(\kappa)
\end{align*}
That is, the utility of a solver $s$ using a step utility function with parameter $\kappa$ is $F_s(\kappa)$, the value of the CDF at $\kappa$. This implies a family of utility functions by which $A$ is preferred to $B$:

\begin{figure}
    \centering
    \includegraphics[width=.9\linewidth]{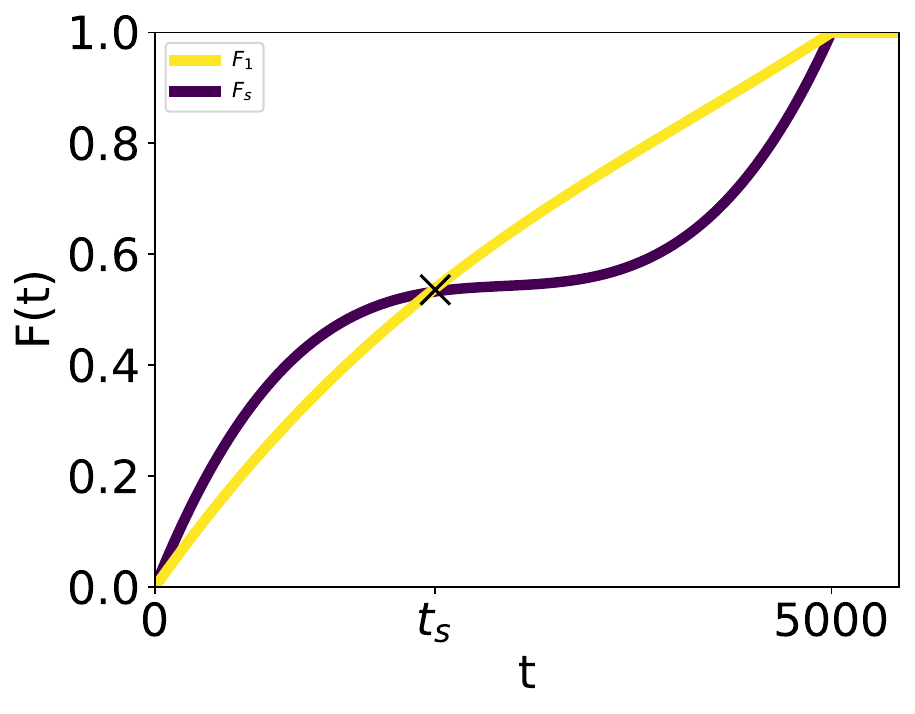}
    \caption{The winning solver is FOSD over solver $s$ over the range $(t_s, 5000)$.}
    \label{fig:cdf_cartoon}
\end{figure}

\begin{quote}
    \emph{If $A$ is FOSD over $B$ in some measurable region $R_A$, then $A$ will be preferred to $B$ for any step utility function that has $\kappa \in R_A$.}
\end{quote}

The point $t = \kappa$ is the only point that matters to us. Whatever interval it falls in is the interval we care about. In fact though, we can see that it does not matter that $u$ is a simple step function. What matters is that $u$ dropped from 1 to 0 within the region $R_A$, where $A$ is FOSD. 

We can rewrite the expression for the utility of a solver $s$ using integration by parts: 
\begin{align*}
    U_s &= \int_{t=0}^\infty u(t) dF_s(t) \\
    &= -\int_{t=0}^\infty F_s(t) du(t) .
\end{align*}
This changes an integral with respect to $F_s(t)$ into an integral with respect to $u(t)$. Note that since $u$ is monotonically decreasing, $du(t)$ is non-positive. Let's define $\mu(t) = 1 - u(t)$ so that $d\mu = - du$ and we have
\begin{align*}
    U_s &= \int_{t=0}^\infty F_s(t) d\mu(t) .
\end{align*}
Now if the utility function is constant outside of the region $R_A$ where solver $A$ is FOSD, then the change in utility function $d\mu(t)$ will be 0 outside of this region, and the utility of any solver $s$ will be 
\begin{align*}
    U_s &= \int_{t \in R_A} F_s(t) d\mu(t) .
\end{align*}
Since solver $A$ is FOSD throughout $R_A$, we will have $F_A(t) \ge F_B(t)$ for all $t \in R_A$, and $F_A(t) > F_B(t)$ for some $t \in R_A$, and thus $U_A > U_B$ for any other solver $B$. We can again infer a broad class of utility functions for which a solver $A$ is preferred:
\begin{quote}
    \emph{If $A$ is FOSD over $B$ in some measureable region $R_A$, then $A$ is preferred to $B$ for any utility function that drops from 1 to 0 entirely within $R_A$.}
\end{quote}

For a general utility function $u$, the drop in $u$ at any point $t$ is $du(t)$. We can be interpret this as a ``utility mass'' that ``measures'' how marginally important runs at $t$ are to us. Since a utility function $u$ is monotonically decreasing from 1 to 0, it is equivalent, under mild conditions, to a survival function (i.e., it is ``one minus a CDF''): $u(t) = 1 - \mu(t)$, where $\mu$ is a CDF. Just as a CDF describes a measure of probability mass, a utility function can be seen as describing a measure of ``utility mass'' (or rather of \emph{loss} of utility mass, since runtime is bad). 

A solver $A$ is preferred when all of $u$'s utility mass falls from 0 to 1 entirely within $R_A$, the region where $A$ is FOSD. In general, any utility function that puts all of its utility mass in $R_A$ will make $A$ preferred. More generally still, $A$ will be preferred to $B$ by any utility function that puts relatively more utility mass in $R_A$ than in $R_B$, the region where $B$ is FOSD. 

\begin{align*}
    U_A - U_B &= \int_{t=0}^\infty \big(F_A(t) - F_B(t)\big) d\mu(t) \\
    &= \int_{t \in R_A} \Big(F_A(t) - F_B(t)\Big) d\mu(t) \\
    &\quad - \int_{t \in R_B} \Big(F_B(t) - F_A(t)\Big) d\mu(t) 
\end{align*}

We can interpret the integral $\int_{t \in R_A} \big(F_A(t) - F_B(t)\big) d\mu(t)$ as the ``size'', according to the ``utility measure'' $\mu$, of the area between the two CDFs over the region $R_A$, where $A$ is FOSD. This is the ``size'' of the difference between the CDFs over the region where the winning solver is FOSD. In effect, this integral measures, according to our utility function, just how FOSD solver $A$ is over solver $B$. Similarly, the integral $\int_{t \in R_B} \big(F_B(t) - F_A(t)\big) d\mu(t)$ ``measures'', according to $\mu$, the ``size'' of the area between the two CDFs over the region $R_B$, where $B$ is FOSD.

\begin{figure}
    \centering
    \includegraphics[width=.9\linewidth]{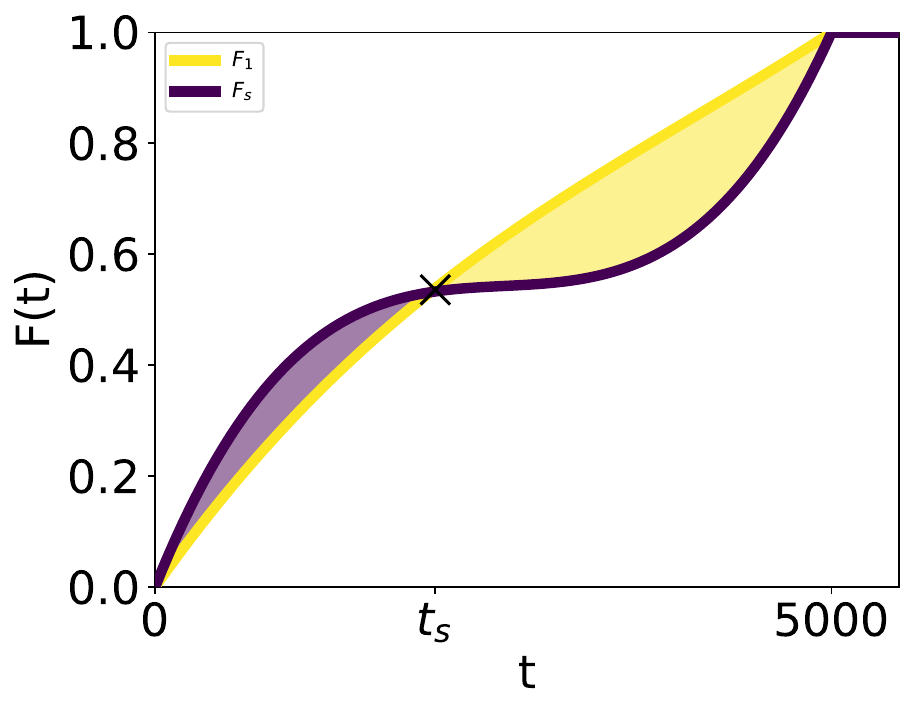}
    \caption{{\small The difference in the ``utility measure'' of the two regions determines which solver is preferred.}}
    \label{fig:cdf_cartoon_shaded}
\end{figure}

This gives us the most general class of utility functions for which a solver $A$ is preferred:
\begin{quote}
    \emph{If $A$ is FOSD over $B$ in some measurable region $R_A$ and $B$ is FOSD over $A$ in some measurable region $R_B$, then $A$ is preferred to $B$ for any utility function that measures the area between the CDFs as being larger in $R_A$ than in $R_B$.}
\end{quote}

We can relate this geometrically to our cartoon plot of the runtime CDFs in \cref{fig:cdf_cartoon_shaded}. If a utility function ``measures'' the yellow region as being bigger than the purple region, then solver $A$ will be preferred. 

This ``utility as a measure'' interpretation is enlightening, and is particularly clear when we are dealing with step utility functions. More generally though, we can simply compute the integrals numerically for different utility functions and see which solver is best.

\subsection{PAR Utility Functions and Beyond}

We can say more about what scoring functions will cause a solver to be ranked highly or poorly. Looking at \cref{fig:rankdiff} we can see that the winning solver ("SBVA-sbva\_cadical") is the same regardless of the value of $c$, and does not change until $\kappa$ is below about 1000, indicating that this is the best solver by a broad range of criteria. But we can be more specific. We can plot the utility of each solver as a function of the parameter(s) of the utility function. We can then look at which solver is best for each value of the parameter(s). 

\begin{figure}    
    \includegraphics[width=.68\linewidth,right]{img/diffk_optimals_utility.pdf}    
    \hfill 
    \centering
    \includegraphics[width=.9\linewidth,right]{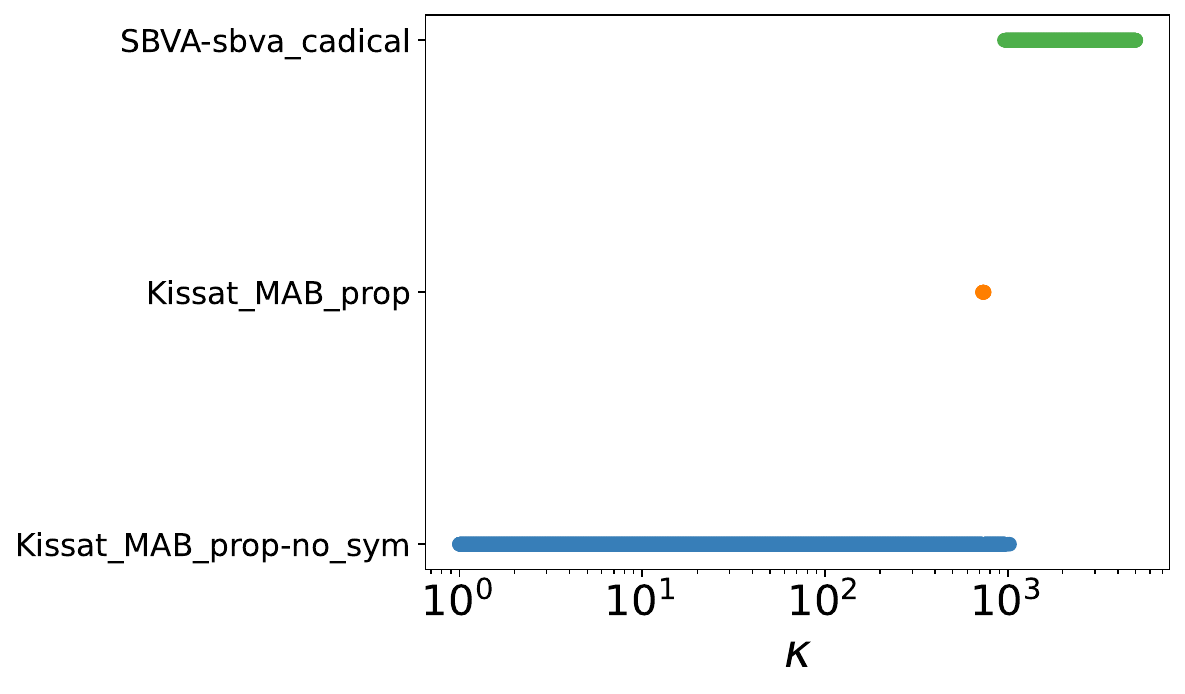}
    \caption{\small Regret according to $PAR(2, \kappa)$ utility function for different values of $\kappa$ and optimal solver as a function of $\kappa$.}

    \label{fig:diffk_optimals_utility_app}
\end{figure}

Holding $c = 2$ fixed, we know that if we are given $\kappa=5000$ seconds to solve our instance then the winning solver is best on average. But what about other values of $\kappa$? \cref{fig:diffk_optimals_utility_app} shows the mean utility of each solver as a function of the chosen parameter $\kappa$. For $\kappa$ above about 1000 the winning solver is best, but for smaller $\kappa$ it is actually the 4th ranked solver (Kissat\_MAB\_prop-no\_sym) that is best. For some $\kappa$ around about 1000 the 3rd ranked solver (Kissat\_MAB\_prop) is best, but all three solvers have very similar performance at this point.  

If we want to choose a solver for some specific application, but we do not know exactly how much time we will be given to solve an instance, we can decide based on whether we think we will be given more or less than 1000 seconds.

\begin{figure}
    \centering
    \includegraphics[width=.72\linewidth,right]{img/diffc_optimals_utility.pdf}    
    \hfill 
    \includegraphics[width=.87\linewidth,right]{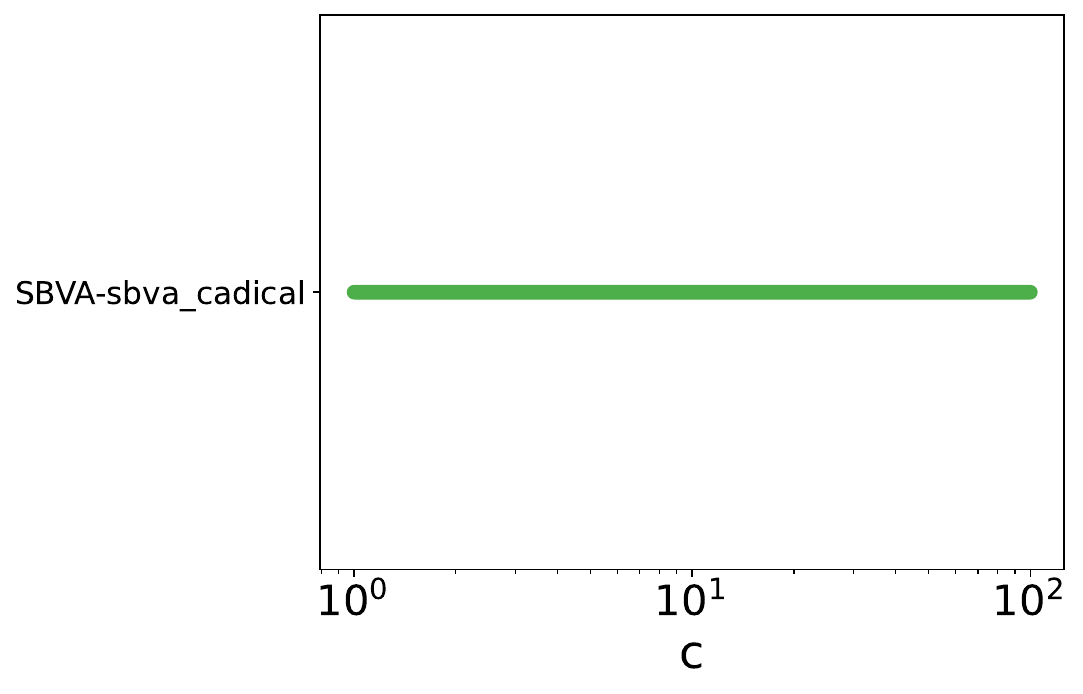}
    \caption{\small Regret using the $PAR(c, 5000)$ utility function for different values of $c$ and optimal solvers as a function of $c$.}
    \label{fig:diffc_optimals_utility_app}
\end{figure}

Holding $\kappa=5000$ fixed, we can ask the same question about the value of $c$. What if we are not exactly sure of the penalty we will face for runs that fail to complete? In this case the situation is particularly simple. \cref{fig:diffc_optimals_utility_app} shows the mean utility for all solvers for different values of $c$, and we can see that the value of $c$ does not affect which solver is ranked first. 

It is natural to consider $PAR(c, \kappa)$ utility functions, since this is how the solvers are ranked in the SAT Competition, but we can look at other utility functions as well. For example, we could use an exponential function with parameter: $u(t) = e^{-\lambda t}$, where $\lambda$ is a parameter. If we know the exact value of $\lambda$ that we want to use, then we can just compute the mean utility for each solver and find which one is best. Otherwise we can plot the mean utility of each solver as a function of $\lambda$ and see how the ranking changes with different values of $\lambda$. 

\begin{figure}
    \centering
    \includegraphics[width=.69\linewidth,right]{img/exp_optimals_utility.pdf}
    \hfill 
    \includegraphics[width=.9\linewidth,right]{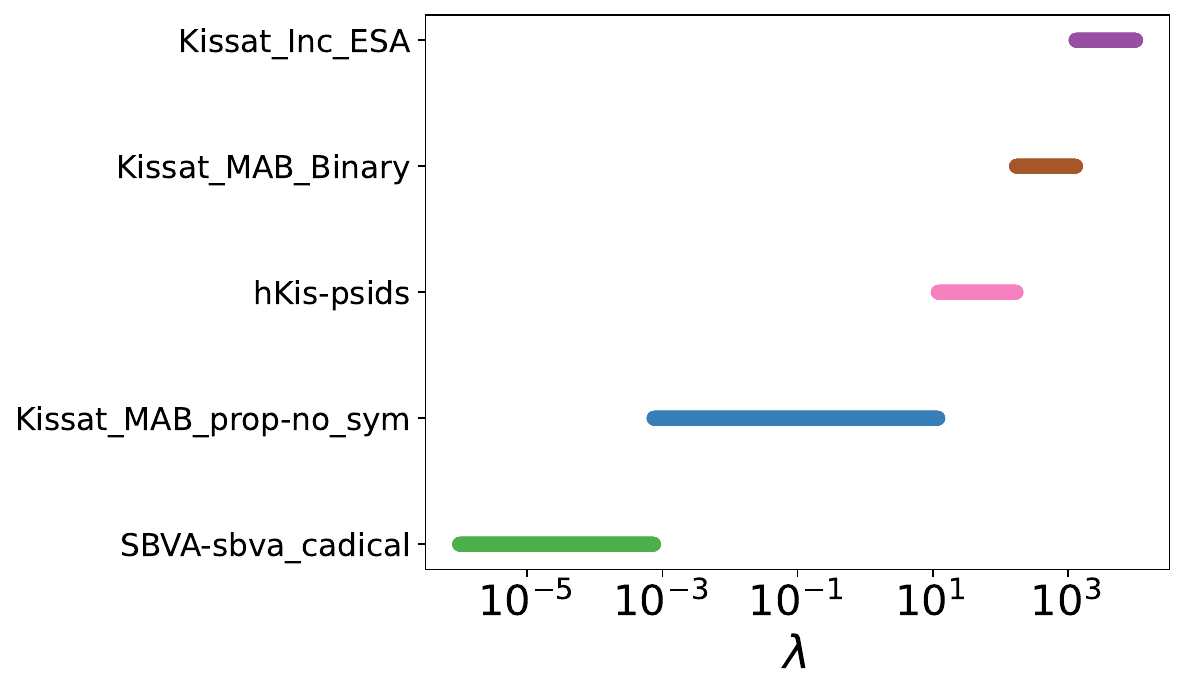}
    \caption{\small Regret using exponential utility function for different values of $\lambda$ and optimal solvers as a function of $\lambda$.}
    \label{fig:exp_optimals_utility_app}
\end{figure}

It is interesting to note that the top-ranked solver (SBVA-sbva\_cadical) and the 4th-ranked solver (Kissat\_MAB\_prop-no\_sym) are best according to the exponential utility function as well for a wide range of parameter values.

\clearpage

\section{Individual Performance Plots}\label{app:plots}

\begin{figure}[h]
    \centering
    \includegraphics[width=.9\linewidth]{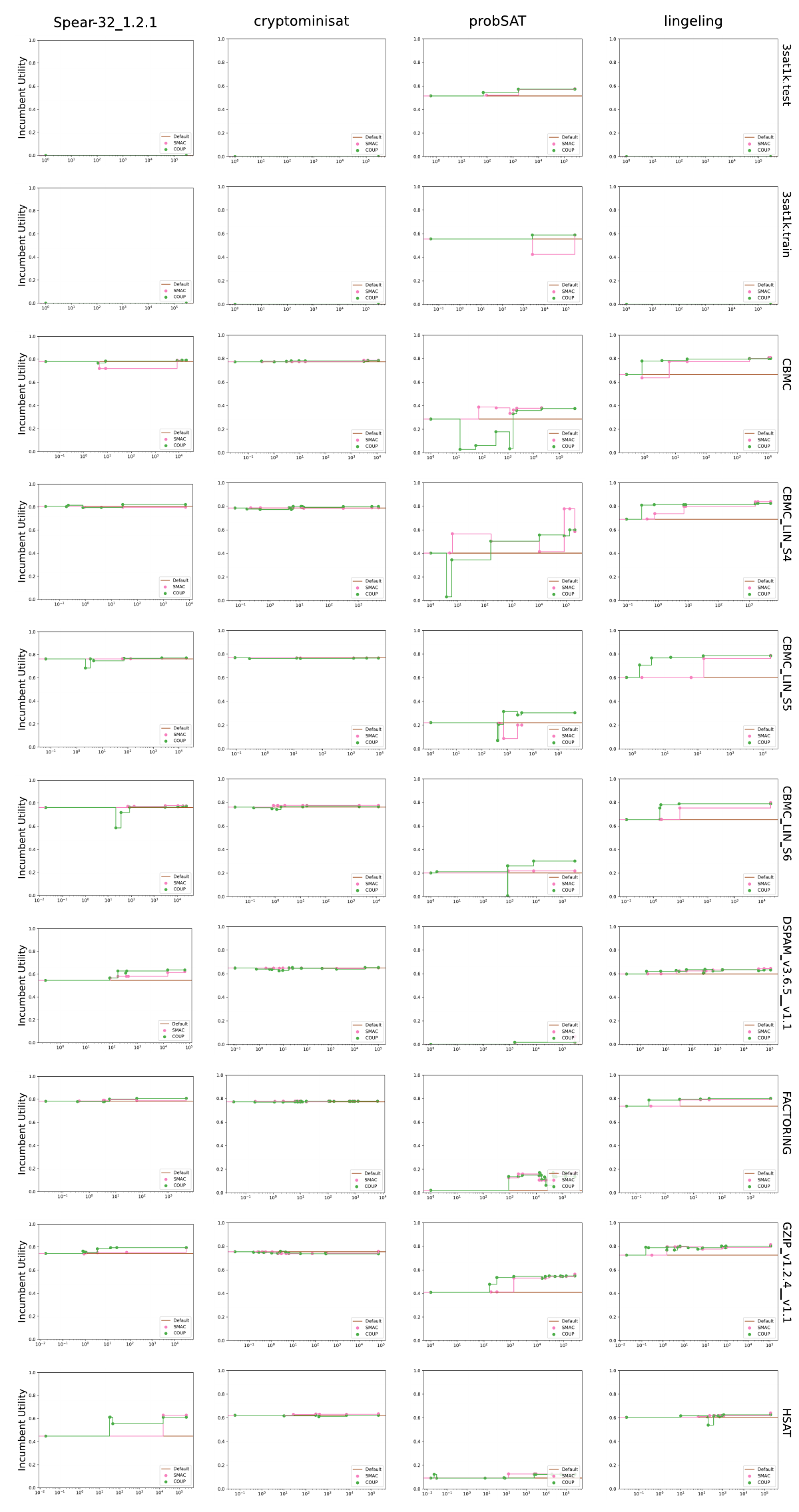}
\end{figure}

\begin{figure}[t]
    \centering
    \includegraphics[width=.9\linewidth]{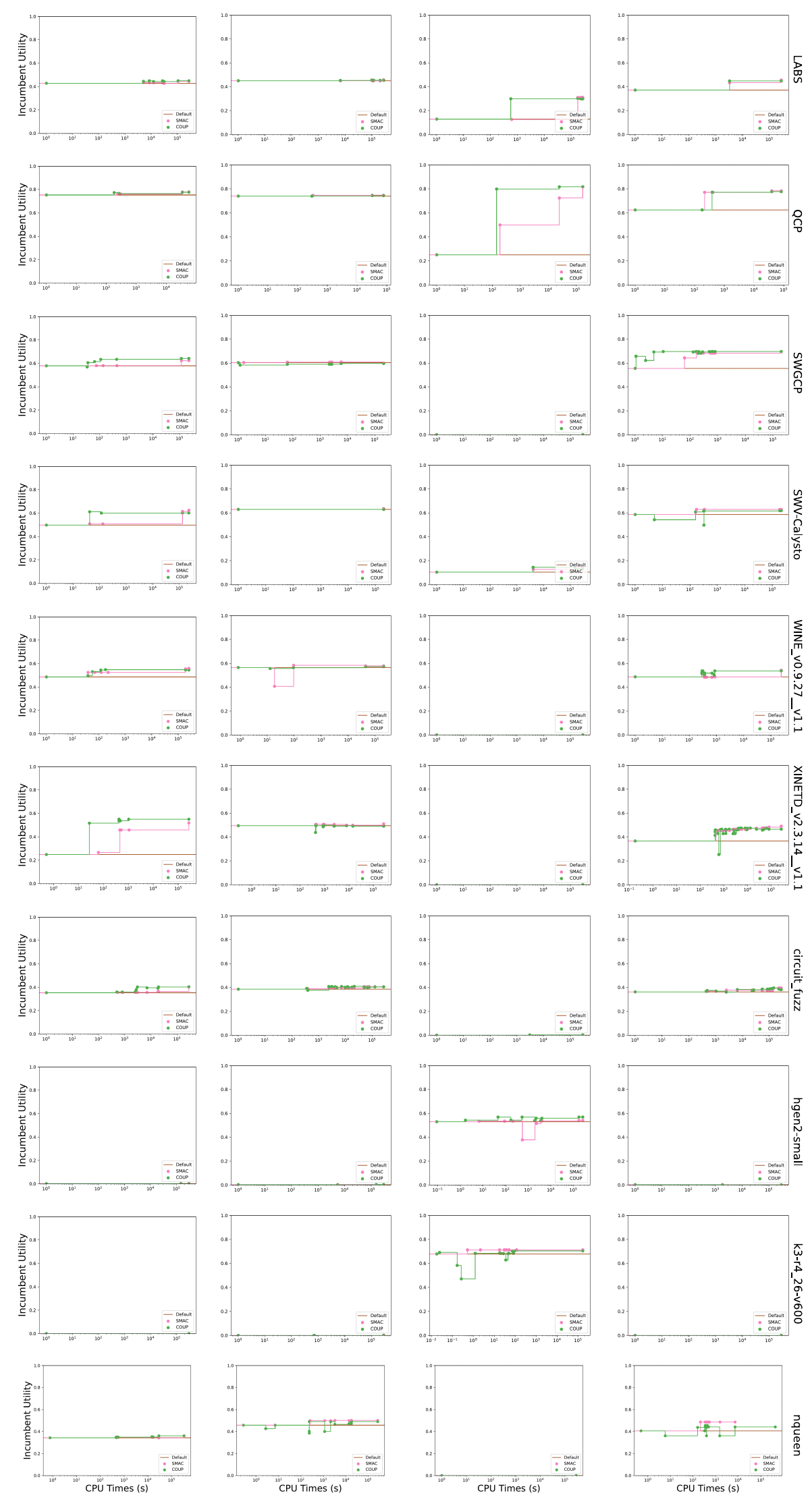}
    \caption{Incumbent utility for COUP and SMAC for different solver--instance set pairs from ACLib.}
\end{figure}

\begin{figure}
    \centering
    \includegraphics[width=\linewidth]{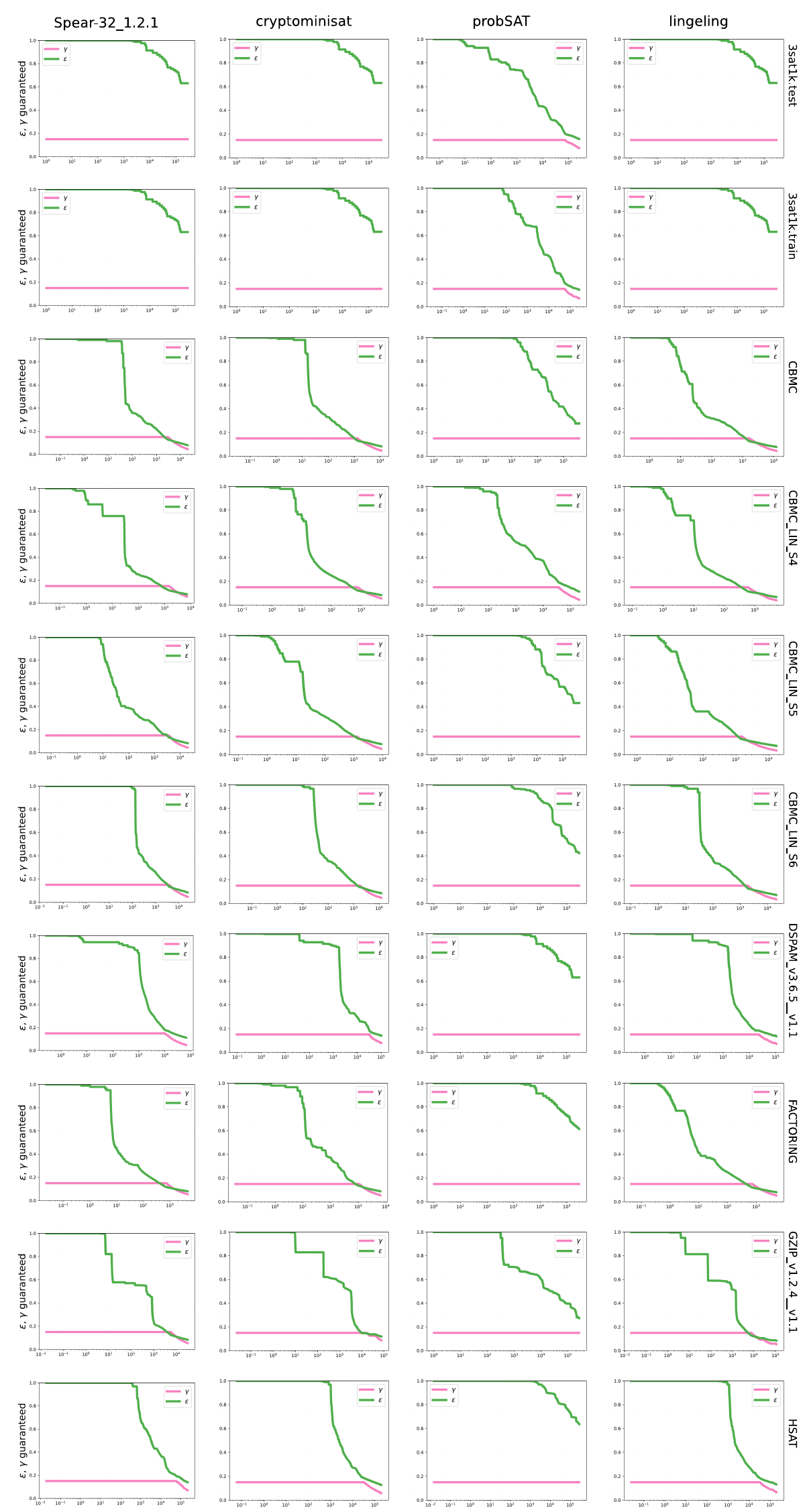}
\end{figure}

\begin{figure}
    \centering
    \includegraphics[width=\linewidth]{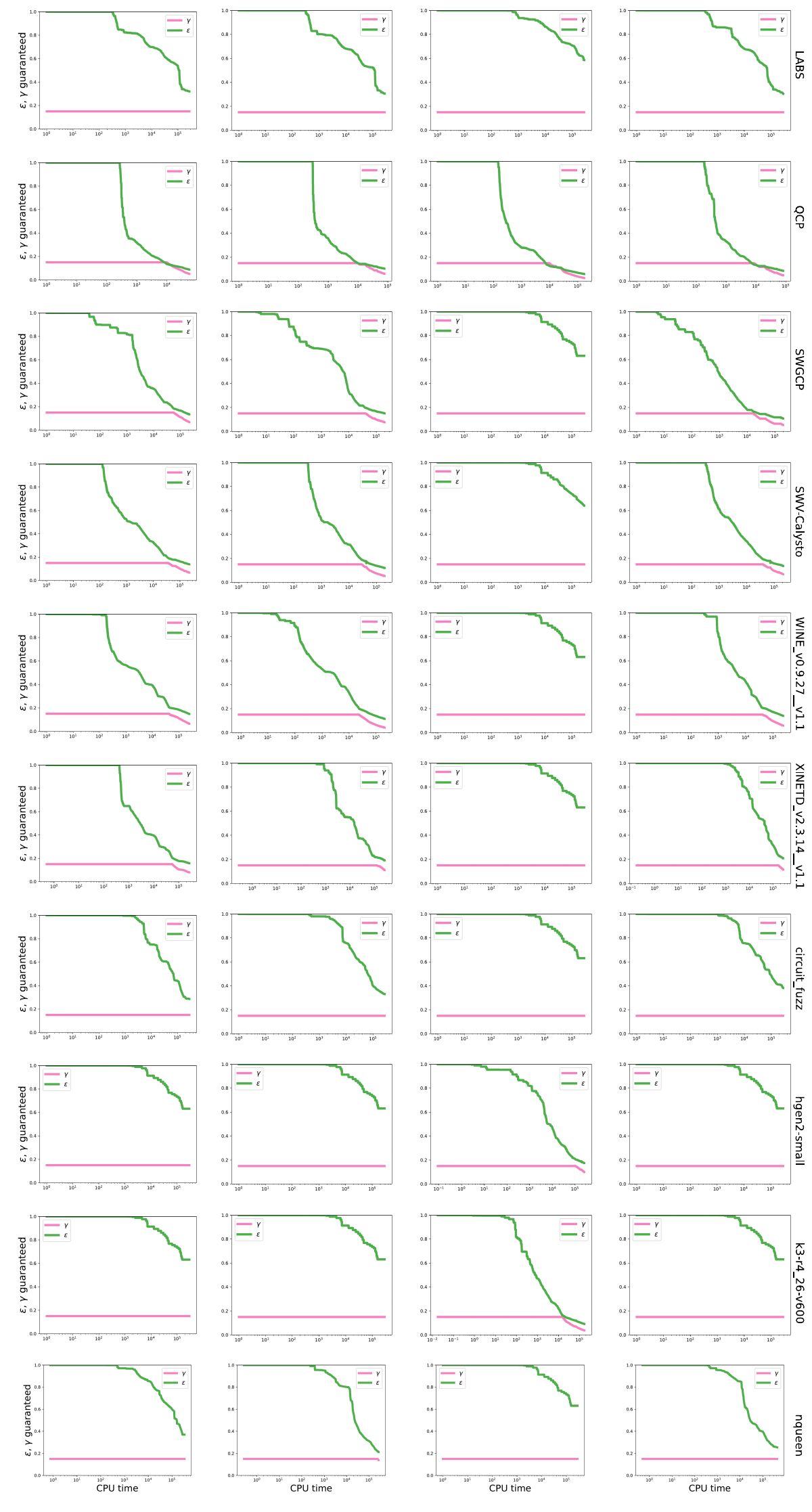}
    \caption{$\epsilon$ and $\gamma$ proved by COUP for different solver--instance set pairs from ACLib.}
\end{figure}

\clearpage

\end{document}